\documentclass[lettersize,journal]{IEEEtran}
\usepackage{amsmath,amsfonts}
\usepackage{algorithmic}
\usepackage{algorithm}
\usepackage{array}
\usepackage[caption=false,font=normalsize,labelfont=sf,textfont=sf]{subfig}
\usepackage{textcomp}
\usepackage{stfloats}
\usepackage{url}
\usepackage{verbatim}
\usepackage{graphicx}
\usepackage{cite}
\usepackage[dvipsnames]{xcolor}
\usepackage{paralist }
\usepackage{booktabs}     
\usepackage{threeparttable}
\usepackage{multirow} 
\newcommand{\weikai}[1]{\textcolor{black}{#1}}
\newcommand{\lizhen}[1]{\textcolor{black}{#1}}
\newcommand{\jwrevision}[1]{\textcolor{black}{#1}}
\newcommand{\shixia}[1]{\textcolor{black}{#1}}
\newcommand{\lizhenrev}[1]{\textcolor{black}{#1}}
\newcommand{\jwrev}[1]{\textcolor{black}{#1}}
\def \etal {{\emph{et al}.\thinspace}}
\def \eg {{\emph{e.g}.\thinspace}}
\def \ie {{\emph{i.e}.\thinspace}}
\newcommand{\system}{{RuleExplorer}}
\begin{document}

\title{\system: A Scalable Matrix Visualization \\for Understanding Tree Ensemble Classifiers}

\author{Zhen Li, Weikai Yang, Jun Yuan, Jing Wu, Changjian Chen, Yao Ming, Fan Yang, Hui Zhang, Shixia Liu}

\markboth{Journal of \LaTeX\ Class Files,~Vol.~14, No.~8, August~2021}%
{Shell \MakeLowercase{\textit{et al.}}: A Sample Article Using IEEEtran.cls for IEEE Journals}


\maketitle

\begin{abstract}
The high performance of tree ensemble classifiers benefits from a large set of rules, which, in turn, makes the models hard to understand.
To improve interpretability,
existing methods extract a subset of rules for approximation using model reduction techniques.
However, by focusing on the reduced rule set, these methods often lose fidelity and ignore anomalous rules that, despite their infrequency, play crucial roles in real-world applications. 
\jwrevision{This paper introduces a scalable visual analysis method to explain tree ensemble classifiers \weikai{that contain tens of thousands of rules.}} 
The key idea is to address the issue of losing fidelity by adaptively organizing the rules as a hierarchy rather than reducing them.
To ensure the inclusion of anomalous rules, we develop an anomaly-biased model reduction method to prioritize these rules at each hierarchical level.
Synergized with this hierarchical organization of rules, we develop a matrix-based hierarchical visualization to support exploration at different levels of detail.
Our quantitative experiments and case studies demonstrate how our method fosters a deeper understanding of both common and anomalous rules, thereby enhancing interpretability without sacrificing comprehensiveness. 
\end{abstract}


\section{Introduction}
\begin{figure}[t]
\centering
\includegraphics[width=1\linewidth]{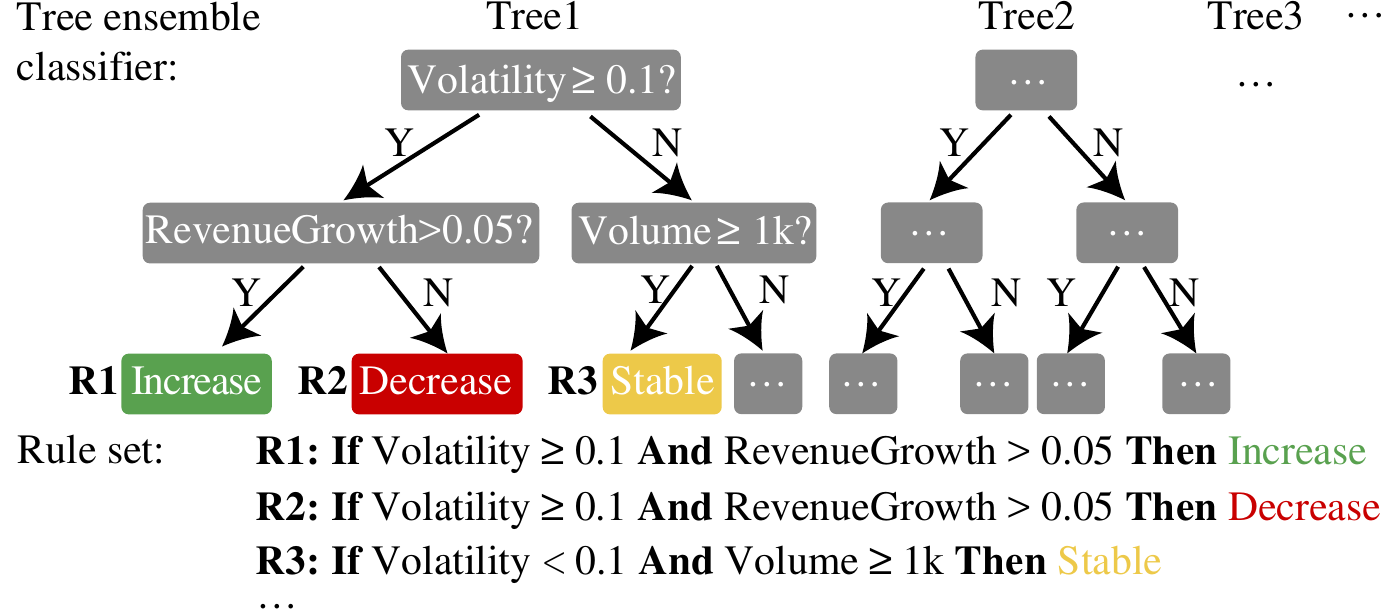}
\vspace{-6mm}
\caption{An illustration of a tree ensemble classifier \lizhenrev{used to predict stock price movement with labels ``increase'', ``decrease'', and ``stable.''}
}
\vspace{-3mm}
\label{fig:tree_ensemble}
\end{figure}

Tree ensemble classifiers, such as random forests and boosted trees, are popular machine learning models ~\cite{hara2018making,liu2018visual,zhou2019deep}.
These models make \weikai{multi-class} predictions using a large number of rules \weikai{(Fig.~\ref{fig:tree_ensemble})}, often exceeding tens of thousands~\cite{hara2018making}.
On the one hand, the large set of rules makes these models powerful and has led them to become winning solutions for many machine learning competitions~\cite{zhou2020kdd}. 
On the other hand, \shixia{while individual rules are simple and} easily understandable, the overall interpretability of these models is compromised by the overwhelming number of rules. 
Therefore, there is a trade-off between the performance and interpretability in these tree ensemble classifiers.
Although performance is the main goal in machine learning competitions, the ability \shixia{of} domain experts to understand and explain decisions is crucial in high-stakes fields such as healthcare, law enforcement, and financial forecasting.\looseness=-1

Several works have been proposed to enhance the interpretability of tree ensemble classifiers 
~\cite{zhao2018iforest,neto2020explainable}. 
However, the scalability issue becomes the main challenge when managing rule sets that extend to tens of thousands or even more. 
A widely adopted strategy to address the scalability issue of a complex model is model reduction~\cite{du2019techniques}, which simplifies the original model by using fewer rules to provide better interpretability.
However, there are two technical challenges in using model reduction techniques directly.

First, in order to explain the overall behavior of the original model with fewer rules, model reduction methods usually focus on extracting common rules that cover most of the samples while ignoring less frequent, anomalous ones.
However, in real-world applications, these anomalous rules often expose potential flaws in the model and thus require further diagnosis \jwrevision{for a complete understanding of the model behavior~\cite{leman2008exceptional}.
}
For example, a credit card application may be approved based on seemingly unrelated conditions, such as having a driver's license.
However, the actual reason is that other applicants who meet these conditions typically have a good credit history and a certain income level.
Previous research has shown that anomaly-biased sample selection methods \shixia{effectively preserve} the overall distribution and highlight anomalous samples for analysis ~\cite{xiang2019interactive,yang2022diagnosing}.
\shixia{As a result, a promising solution is} an anomaly-biased model reduction method that preserves both common rules and anomalous rules.

Second, simply using model reduction inevitably results in a loss of fidelity to the original model. This causes discrepancies between the behavior of the reduced model and that of the original one.
Therefore, only examining the reduced rule set after model reduction may lead to problems in analyzing the whole model. 
For example, an abnormal behavior identified in the reduced model may be difficult to trace back to rules in the original model without access to the whole rule set.
To address this problem, it is essential to find an effective way to examine all the rules of the original model.
Hierarchical visualization has been \jwrevision{shown} 
effective in the exploration of large-scale data~\cite{munzner2014visualization}. 
By re-organizing rather than discarding the rules that are not preserved in model reduction, the issue of losing fidelity can be effectively mitigated.
\shixia{However, the pre-built hierarchies commonly used in 
hierarchical visualization methods~\cite{lyu2024visualization, chen2024enhancing, xiang2019interactive, yuan2022visual,chen2022towards} lack the 
flexibility to accommodate different user preferences during exploratory analysis.
This flexibility is often needed in real-world applications.
For example, in the credit card approval process, \shixia{users frequently} examine different subsets of rules 
by constraining \jwrevision{the values of different attributes (e.g., \emph{Income}, \emph{Years Employed}, etc.) in different ranges.}
As a result, there is a need to 
\jwrevision{flexibly} build the hierarchy for different analysis needs}.

Based on the above analysis, we develop {\system}, a scalable visual analysis tool designed to support interactive 
\shixia{
analysis} of rule sets extracted from tree ensemble classifiers. 
\weikai{The scalability to manage tens of thousands of rules is achieved by} combining anomaly-biased model reduction and hierarchical visualization.
Starting from all the rules, {\system} initially utilizes 
model reduction 
to extract the rules for the first level \jwrevision{in the hierarchy}. 
Subsequently, it dynamically builds the hierarchy in a top-down manner based on user selections.
At every level, a balance between representativeness and diversity is maintained by preserving both common and anomalous rules. 
Synergized with this dynamic rule \jwrevision{hierarchy} organization, a matrix-based hierarchical visualization is developed to support the exploration of rules at different detail levels.
In this visualization, rows and columns correspond to rules and attributes, respectively.
The confidence, coverage, and anomaly scores of each rule are also displayed.
This setup allows users to quickly understand the common behavior of the model and identify 
rules that exhibit abnormal behavior.
Finally, users can zoom in to explore more rules with similar behavior.
Users can also examine the samples associated with certain rules to diagnose potential flaws in the model.
The coordinated analysis between rules and samples further enhances the understanding of model behavior.\looseness=-1

The effectiveness of the anomaly-biased model reduction is demonstrated through quantitative evaluation.
Two case studies further demonstrate the effectiveness of {\system} in analyzing the decision-making process of random forest models and gradient boosted tree models. \lizhen{Note that while we focus on tree ensemble classifiers, the methods are also applicable to other models through model surrogate techniques.}
In summary, the main contributions of this work are:
\begin{compactitem}
\item\noindent{An anomaly-biased model reduction method that preserves both common and anomalous rules.} 
\item\noindent{A matrix-based and dynamically constructed hierarchical visualization that supports 
the exploration of a large-scale rule set at different detail levels.}
\item\noindent{A visual analysis tool that facilitates the understanding, exploration, and validation of tens of thousands of rules.
}
\end{compactitem}

\section{Related Work}

\begin{figure*}[t]
  \centering
    \includegraphics[width=1\linewidth]{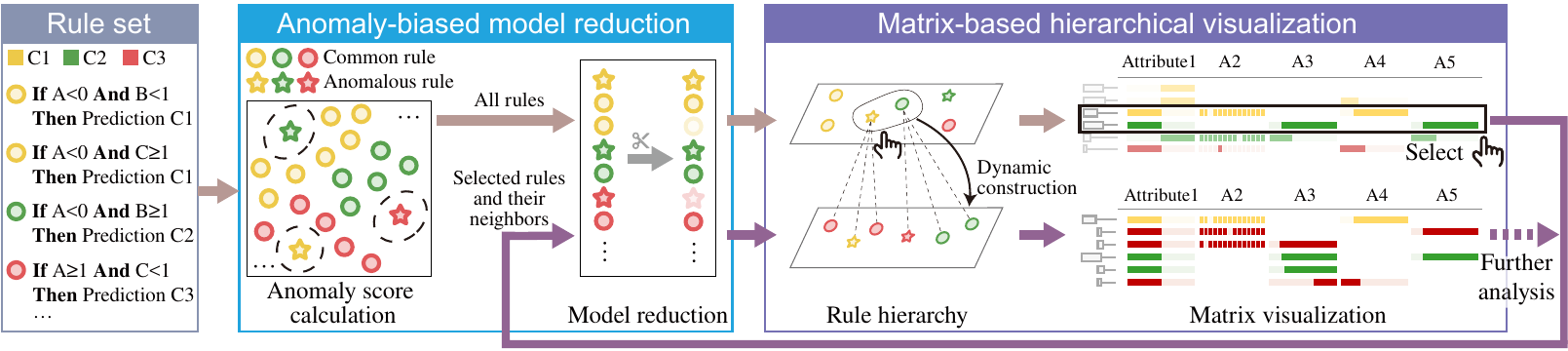}
        \caption{{\system} overview: Given a set of rules from a tree ensemble classifier, the anomaly-biased model reduction calculates the anomaly score for each rule and then extracts the representative rules. \lizhen{Next}, the representative rules are fed into the matrix-based hierarchical visualization for exploration and analysis, where a rule hierarchy is dynamically built based on user selections.
        The brown color indicates the analysis of the representative rules at the top level, and the purple color indicates the iterative analysis subsequently.}
    \label{fig:analysis_pipeline}
\end{figure*}

\subsection{Model Surrogate}
Our model reduction method relates to a category of explanation methods that utilize surrogate models.
These methods use a simpler and more interpretable model to approximate and explain complex models~\cite{craven1995extracting, ribeiro2016lime, tan2018distill, ming2018rulematrix, yuan2024visual, chatzimparmpas2024deforestvis}.

Most model surrogate methods are model-agnostic.
These methods derive a surrogate model from any given machine learning model by approximating the input-output relationship of the model. 
While these methods have good generalizability, they inevitably \lizhenrev{neglect the inner decision logic in tree ensemble models,}  such as node splitting and majority voting.
As a result, these methods cannot precisely reveal how predictions are made. 

For tree ensemble models, the surrogate methods proposed in
\cite{meinshausen2010node, vidal2020born} leverage the characteristics of tree ensembles for more accurate approximations. 
Meinshausen~\cite{meinshausen2010node} selected representative rules from a random forest by solving a quadratic programming problem with linear inequality constraints.
Vidal~\etal\cite{vidal2020born} constructed an equivalent single decision tree from a given tree ensemble 
to improve model interpretability.
Although these methods successfully preserve common rules that well summarize the behavior of the original model, they \lizhenrev{inevitably lose fidelity because other rules are discarded, especially those} anomalous rules that are important for revealing potential flaws in the model.
\weikai{In contrast, our method} re-organizes a tree ensemble into a hierarchy of rules to mitigate the issues of losing fidelity, and uses anomaly-biased model reduction to maintain a balance between the common rules and the anomalous rules at each hierarchical level. 

\subsection{Visualization for Model Analysis}
The visualization community has developed a number of methods and systems to visually interpret, analyze, and diagnose machine learning models. 
Initial efforts and their subsequent variations aim to illustrate a variety of performance measures such as precision, recall, and between-class confusion~\cite{ren2016squares,gortler2021neo,chen2024unified}.
Most recent efforts focus on explaining the inner workings of a machine learning model based on architecture exploration and the reasoning of its prediction.

Architecture visualization methods focus on explaining the inner workings of the model by analyzing specific components in the model~\cite{yang2024foundation}, such as neurons and kernels in convolutional neural networks~\cite{Tzeng2005, liu2017towards, bilal2017convolutional}, hidden states in recurrent neural networks~\cite{ming2017understanding, strobelt2017lstmvis, cashman2018rnnbow, kwon2018retainvis, sawatzky2019visualizing}, attention heads in transformers and other attention-based models~\cite{strobelt2018s, derose2021attention}, or tree structures in decision trees~\cite{van2011baobabview, muhlbacher2017treepod, liu2018visual}.
These visualizations are designed to help \lizhen{model developers} analyze and diagnose machine learning models.
However, for domain experts who have little knowledge of machine learning, such as financial analysts or doctors, the complex workflows and model-architecture-specific designs are of little help for their understanding and utilizing the models.
To \lizhen{support} these domain experts, our work focuses on explaining rules learned from tree ensemble classifiers, as rules are more understandable to them. 
If offered the right tool, they can make sense of the learned rules and validate the rules of interest.

\subsection{Rule Visualization}
Existing efforts to visualize rules can be \jwrevision{categorized as single-rule-oriented or multiple-rule-oriented.} 
Single-rule-oriented visualization~\cite{han2000ruleviz, chatzimparmpas2023visruler} focuses on visualizing samples covered by a single rule, while multiple-rule-oriented visualization seeks to visually illustrate interactions and relationships between multiple rules.
Our work aligns with the latter.
Common methods for multiple-rule-oriented visualization include the node-link diagram and the matrix diagram. 
\lizhen{
Other compact representations such as treemaps~\cite{muhlbacher2017treepod}, icicle plots~\cite{eirich2022rfx}, and sunburst diagrams~\cite{wang2022timbertrek} \jwrevision{are also used} to handle a larger number of rules.
While these methods are effective at presenting an overview of the rules, they may not always provide the level of details necessary for an in-depth analysis of individual rules.
\jwrevision{Therefore,} we focus on introducing the first two methods.}

The node-link diagram~\cite{van2011baobabview,tam2016analysis,muhlbacher2017treepod,jia2020visualizing} naturally represents the decision tree structure or summarizes multiple rules by aggregating the same conditional statement.
For example, BaobabView~\cite{van2011baobabview} employs a node-link diagram to visualize the rules of decision trees and further encodes the split points and data flows in the tree.
\lizhen{TreePOD~\cite{muhlbacher2017treepod} extends this design by utilizing a treemap to display the distribution of samples covered by different tree nodes.}
SuRE~\cite{yuan2024visual} summarizes the entire random forest into a node-link diagram while also retaining useful logical relationships between different rules.
While the node-link diagrams effectively illustrate relationships between rules, they struggle to remain clear and manageable when displaying tens of thousands of rules.
The matrix diagram~\cite{ming2018rulematrix, neto2020explainable, yuan2021exploration} is a natural graphical representation of a list of rules.
RuleMatrix~\cite{ming2018rulematrix} visualizes a falling rule list using a matrix diagram with rows as rules, columns as attributes, and cells as conditions using histogram encoding. 
\lizhenrev{DeforestVis~\cite{chatzimparmpas2024deforestvis} provides attribute-based explanations using simplified decision trees, allowing users to incrementally generate them to explore the trade-off between complexity and fidelity.}
ExplainableMatrix~\cite{neto2020explainable} also employs a matrix diagram, but distinguishes itself by using colored rectangles in matrix cells to encode the conditions for different classes.
However, these methods show all the rules in one view without filtering, and still encounter the scalability issue in actual scenarios.

To address the \jwrevision{scalability issue,} we build a matrix-based hierarchical visualization to show representative rules at different levels to help practitioners understand, explore, and validate tree ensemble classifiers.

\section{Design of {\system}}
\label{sec:overview}
\subsection{Design Goals}

We designed {\system} in collaboration with eight domain experts (E$_1$ -- E$_8$) from banks and quantitative trading companies. 
They utilize tree ensemble classifiers for credit card approval and stock trading.
All of them are not co-authors of this work. 
We distilled the design requirements based on discussions with these domain experts and a literature review.
\looseness=-1

\noindent \textbf {R1. Hierarchically organize large-scale rule sets.
}
In real-world applications, a tree ensemble classifier often contains tens of thousands of rules, which makes it impractical to examine each rule individually.
Although model reduction methods can extract representative rules to explain model behavior, they inevitably lose fidelity~\cite{meinshausen2010node, ming2018rulematrix, yuan2024visual}.
Domain experts have expressed the need to explore the entire rule set for analysis.
Therefore, reorganizing rather than reducing the rules is preferred. A hierarchical reorganization is a promising solution 
due to its effectiveness in the analysis of large-scale data~\cite{shneiderman2003eyes}.
Moreover, during the reorganization, both common and anomalous rules should be preserved \lizhen{at each level} to ensure a comprehensive understanding and diagnosis of potential problems.\looseness=-1

\textit {R1.1 Preserve common rules to enable the understanding of the overall model behavior}.
All the experts emphasized the importance of examining the common rules learned by the model to establish trust.
As E$_2$ said, once he understood the common rules used by the model in credit card approval, he could trust the model predictions.
This would relieve his burden of examining each application individually.
In addition, E$_3$ expressed concerns that model errors could result in financial losses. Thus, he would like to open the ``black box'' model and understand the overall model behavior from the common rules before trusting the decisions made by the model.\looseness=-1

\textit {R1.2 Preserve anomalous rules to enable the diagnosis of potential model flaws}.
Anomalous rules indicate abnormal model behavior deviating from the common ones, which reflect potential model flaws.
E$_2$ stated, ``Anomalous rules usually cover only a small number of samples, which may be problematic samples and deserve further analysis.''
E$_1$ was also interested in examining anomalous rules, ``If a rule approves applications, but with conditions suggesting low repayment ability and instability, it is likely flawed.''
All the experts believed that examining these rules helps uncover blind spots in the model and opens up opportunities for model improvement.\looseness=-1

\noindent \textbf {R2. Effectively explore and analyze the rules.}
The analysis of a large-scale rule set requires efficiently identifying the rules of interest and analyzing them in context.
This leads to \weikai{three} major requirements.
\weikai{First, an intuitive and informative overview of the rule set is essential to help users quickly gain insights and start their analysis.}
\weikai{Second, support for dynamic hierarchical exploration is necessary to enable detailed navigation through various levels of detail and facilitate the analysis of different rule subsets.
Third,} rich interactions should be provided to allow for an in-depth analysis of rules concerning their utilized attributes and covered samples.
To this end, a scalable and flexible interactive environment is crucial for effective exploration and analysis of rules.

\lizhen{
\textit {R2.1 Present rule sets using an intuitive and informative visual representation.}
\jwrevision{To effectively visualize a hierarchically organized rule set, a requirement is to both reveal the hierarchical structure and present detailed information on attributes within the rules.
The commonly used node-link-based~\cite{yang2021interactive} or scatterplot-based~\cite{xiang2019interactive} hierarchical visualizations focus on the former but require additional interactions to access detailed information, which is not ideal for analyzing rule sets. 
\weikai{For example, E$_1$ expressed a desire for a quick overview that allows immediate identification of important rules and attributes.}
Moreover, simple and intuitive visual representations are always preferred because they facilitate quicker comprehension and reduce cognitive load.} 
\jwrevision{All these call} for visual representations that are both intuitive and informative \weikai{to meet the needs of domain experts.}}

\textit {R2.2 Cater to different analysis preferences.}
Users may 
have different analysis \weikai{needs} and 
focus on different subsets of rules 
during the exploration.
For example, E$_7$ mentioned that in the overall analysis of stocks, he would pay more attention to the attributes directly related to price changes, \weikai{such as moving average price and price volatility}.
\weikai{When he focused on low-cap stocks, he would like to examine} the attributes related to \weikai{stock liquidity, such as turnover ratio and market depth}.
With \jwrevision{the} different analysis 
\weikai{needs}, he 
\jwrevision{zoomed} into different subsets of rules.
Compared to a static hierarchy, which organizes the rules with a pre-defined criterion, a dynamic hierarchy is better at catering to users' different preferences~\cite{elmqvist2009hierarchical, chen2021oodanalyzer, zhou2023cluster}.

\textit {R2.3 Locate rules of interest and relate them to training samples.}
When exploring the entire rule set or specific rule subsets, experts expressed their interest in analyzing rules based on various properties.
E$_1$ was particularly interested in how the model would handle the borderline cases in the credit card application.
Thus, he wanted to examine the low-confidence rules that were mostly relevant to these cases.
To facilitate this, we provide a set of interactions, such as sorting rules by different properties, to help identify the rules of interest. Additionally, 
domain experts emphasized the importance of knowing the samples from which a rule was learned.
For example, before making investment decisions based on some rules, E$_7$ wanted to validate these rules by examining the training samples that contributed to the learning of these rules. This highlights the need to relate rules to training samples.

\subsection{System Overview}

Guided by the aforementioned design requirements, we developed {\system} to help domain experts explore and analyze the large-scale rule sets in tree ensemble classifiers. 
As illustrated in Fig.~\ref{fig:analysis_pipeline}, {\system} 
consists of two modules: anomaly-biased model reduction and matrix-based hierarchical visualization.
The anomaly-biased model reduction module first calculates an anomaly score for each rule. 
Based on the anomaly scores, it then extracts representative rules through model reduction for subsequent visualization. 
The extracted representative rules not only approximate the behavior of the input rule set but also preserve anomalous rules. 
The matrix-based visualization module visualizes the representative rules in a matrix and supports users in exploring the rules by dynamically constructing the rule hierarchy based on user selections. More specifically, users can select some rules of interest in the matrix for a more detailed examination. 
Upon selection, the selected rules and their neighborhood are input into model reduction to extract the representative rules at the next level in the hierarchy.
The matrix visualization is updated accordingly with these new representative rules. 
Users can continue to select and analyze these rules in a similar way. 

\section{Anomaly-Biased Model Reduction}
\label{sec:reduction}
\lizhen{The developed anomaly-biased model reduction method aims to 
\jwrevision{enable both} a comprehensive explanation of the overall model behavior and the identification of potential flaws (\textbf{R1.1}, \textbf{R1.2}).}
\lizhen{To this end}, we first employ logistic regression to calculate the anomaly scores for all the rules in the model.
Then, we formulate the anomaly-biased model reduction as a subset selection problem that extracts the rules to approximate the behavior of the given rule set on their covered samples, while prioritizing the anomalous rules with high scores.

\subsection{Anomaly Score Calculation}
To identify the anomalous rules, we need to calculate an anomaly score for each rule.
This is achieved by converting the rules into feature representations and then employing logistic regression for anomaly detection.

\noindent \textbf{\lizhen{\jwrevision{Rule representation.}
}}
In a tree ensemble classifier, each rule is represented by the conjunction of \jwrevision{conditions} 
on attributes and a corresponding prediction.
\lizhen{Inspired by Zhao~\etal\cite{zhao2018iforest}, we vectorize the condition on each attribute and concatenate them to obtain the feature representation of a rule. }
There are two types of attributes.
First, the condition on a categorical attribute is represented by a vector indicating the distribution of the categories that a sample can fall into.
For example, assume that there are three categories $F_1$, $F_2$, and $F_3$ for a categorical attribute $F$, with $10$, $20$, and $30$ samples in each category, respectively.
Then, a condition $F \in \{F_1, F_2\}$ can be represented by $[10 / (10+20+30), 20 / (10+20+30), 0] = [1/6, 2/6, 0]$.
Second, the condition on a numerical attribute is represented by a two-dimensional vector indicating the 
lower and upper bounds of the \lizhen{value range that satisfies the condition}. 
\jwrevision{To enable a fair comparison between conditions on different attributes with varying scales and distributions, the quantile normalization~\cite{bolstad2003comparison} is used to transform all the attribute distributions  to the uniform distribution in the range $[0,1]$.} 

\noindent \textbf{\lizhen{Anomaly score \jwrevision{calculation}.
}} With the feature representations and the corresponding predictions of the rules, we fit a logistic regression model.
With this model, given a rule $r_j$, we obtain its probability $p_j$ leading to its corresponding prediction.
Accordingly, we calculate its anomaly score as $s_j = 1 - p_j$, which measures the deviation of $r_j$ from the majority of rules~\cite{nurunnabi2012outlier}.
As anomalous rules exhibit behavior deviating from the common rules, they have higher anomaly scores.


\subsection{Model Reduction}
\jwrevision{Model reduction} can be formulated as a subset selection. 
Given a rule set $R = \{r_j\}_{j=1}^m$ associated with their anomaly scores $\{s_j\}_{j=1}^m$ and a set of samples $X=\{x_i\}_{i=1}^n$ belonging to \jwrevision{$C\ge 2$ classes}, 
our method extracts a subset of representative rules $R^{*}$ from $R$ by solving the \jwrevision{optimization problem:} 

\begin{equation} 
\label{eq:loss}
\begin{aligned}
\underset{\{z_j\}_{j=1}^m}{\text{argmin}} 
& \quad \frac{1}{n}\sum_{i=1}^{n}\mathcal L(\hat y_i, \ell_i) - \lambda \frac{1}{M}\sum_{j=1}^{m}z_j s_j, \\
\text{s.t.}
& \quad z_j \in \{0,1\},\ \sum_{j=1}^{m} z_j \leq M. 
\end{aligned}
\end{equation}

Here, $\hat y_i = [\hat y_{i,c}|c=1, \cdots, C]$ is a vector of prediction scores of $x_i$ by the extracted rule set $R^{*}$, where $\hat y_{i,c}$ is the prediction score of $x_i$ to be of class $c$. 
$\ell_i$ is the prediction label of $x_i$ by the given rule set $R$.
\weikai{$\mathcal{L}(\hat{y}_i, \ell_i)$ is the loss function that measures the prediction difference between the extracted rule set 
$R^{*}$ and the original \jwrevision{rule set} 
$R$}.
$z_j$ is the indicator of whether the rule $r_j$ belongs to $R^{*}$.
The first term ensures the consistency between the predictions of the given rules and the extracted rules. 
The second term prioritizes the selection of anomalous rules by maximizing the anomaly scores of the extracted rules.
$\lambda$ is a factor to balance the effect of the two terms.
\weikai{$M$ is the maximum size of the extracted rule subset $R^{*}$. }\looseness=-1

\begin{figure}[t]
\centering
\includegraphics[width=0.95\linewidth]{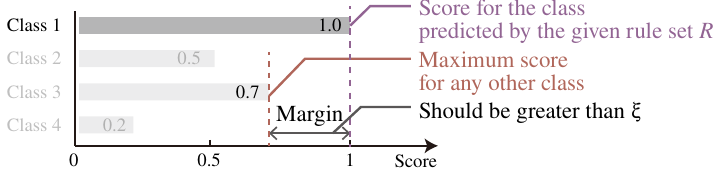}
\caption{\jwrevision{An example illustration of multi-class hinge loss. 
}}
\label{fig:margin}
\end{figure}


Next, we introduce how to calculate $\mathcal{L}(\hat{y}_i, \ell_i)$.
To encourage that $R^{*}$ and $R$ make consistent predictions, the idea is to increase $\hat y_{i,\ell_i}$, \ie, the prediction score of $x_i$ to be of class $\ell_i$, while decreasing other $\hat y_{i, c}$ scores $(c\neq\ell_i)$.
Therefore, we adopt multi-class hinge loss~\cite{crammer2001algorithmic}:

\begin{equation} 
\label{eq:hinge_loss}
\begin{aligned}
\mathcal L(\hat y_i, \ell_i) = \mathop{max}(\xi - (\hat y_{i, \ell_i} - \mathop{max}\limits_{c \neq \ell_i}\ \hat y_{i, c}), 0).
\end{aligned}
\end{equation}

For a specific sample $x_i$, as shown in Fig.~\ref{fig:margin}, the multi-class hinge loss encourages the score for class $\ell_i$ to be greater than any other class by a margin of at least $\xi$.
This helps improve the generalization ability of the extracted rules~\cite{rosasco2004loss}.
\lizhenrev{To reduce the user burden of parameter adjustments and improve ease of use, we determine} $\xi$ and $\lambda$ using a grid search.
\lizhenrev{This search spans values from from 0.1 to 1.0, with a step size of 0.1, effectively balancing} fidelity and anomaly scores.
\lizhenrev{Specifically, we first set $\lambda$ to 0 and searched for $\xi$ that achieved the highest fidelity, then adjusted $\lambda$ to improve anomaly scores while  keeping fidelity loss within 1\%.}
\lizhen{$M$ is set to 80 based on the screen constraints and suggestions from the experts}, \lizhenrev{as they prefer to examine as many rules as possible while maintaining visual clarity for rule analysis.
We also conducted a sensitivity analysis of $M$ in the supplemental materials to justify this choice.}\looseness=-1


Eq.~(\ref{eq:loss}) is an integer linear programming problem, which is NP-hard and time-consuming to solve.
\lizhenrev{Even advanced commercial solvers cannot solve it within reasonable time limits due to the extremely large basis matrix and overwhelming memory consumption resulting from thousands of rules and samples.}
To solve it efficiently, we relax the \jwrevision{$z_j$} 
to continuous variables.
The relaxed problem becomes a linear programming problem and can be solved in polynomial time~\cite{ignizio1994linear}.
Then, we adopt deterministic rounding to discretize \jwrevision{$z_j$} 
and determine which rules belong to $R^{*}$.

\section{Matrix-Based Hierarchical Visualization}

\begin{figure}[t]
\centering
\includegraphics[width=1\linewidth]{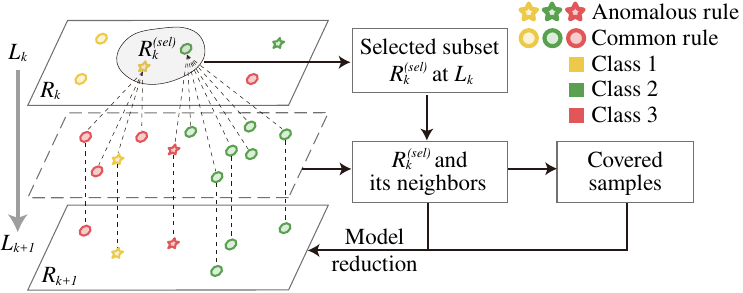}
\caption{Dynamic rule hierarchy construction between consecutive levels.} 
\label{fig:hierarchy}
\end{figure}

\begin{figure*}[t]
\centering
\includegraphics[width=1\linewidth]{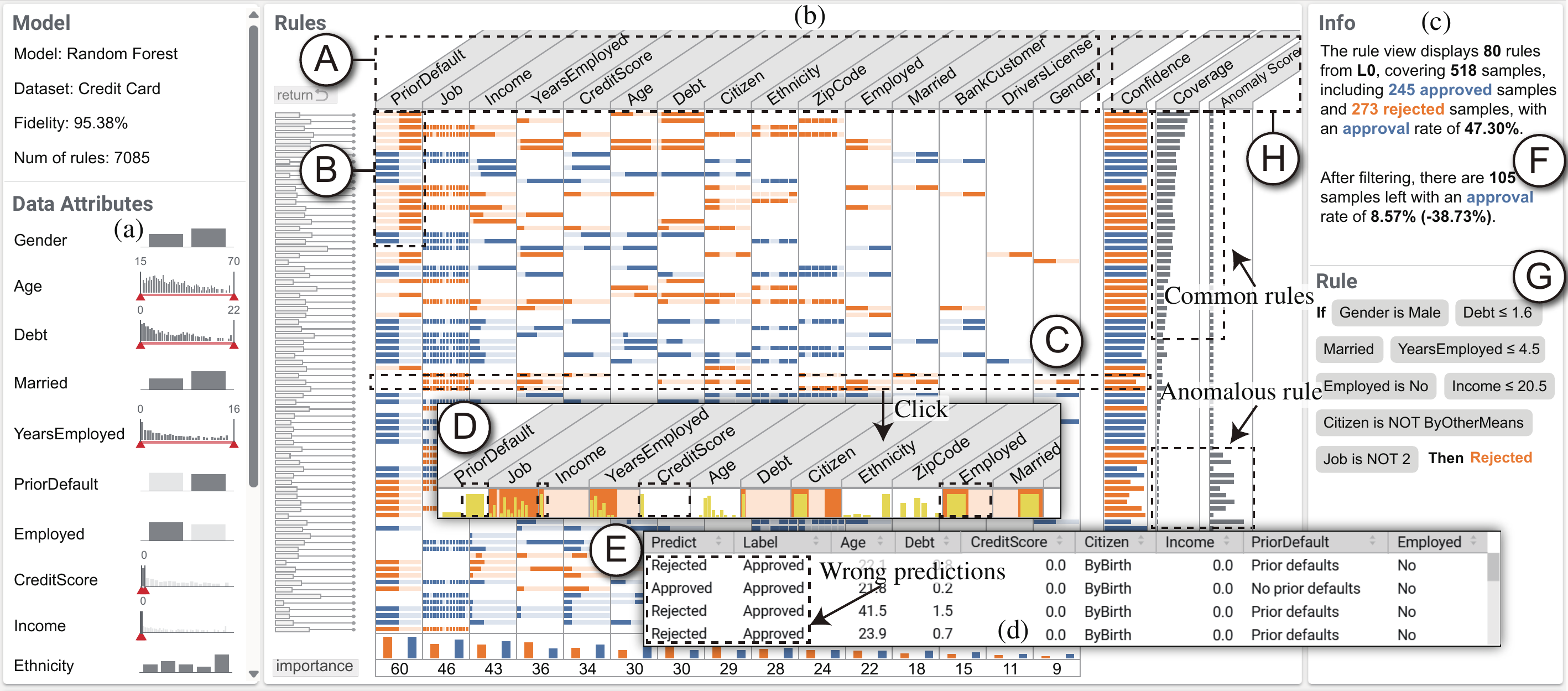}
\caption{{\system}: 
(a) \textit{attribute view} shows the attribute distribution and enables sample filtering; (b) \textit{matrix view} shows the representative rules at a certain level of the rule hierarchy; (c) \textit{info view} shows the overall statistics of the displayed rules and samples; (d) \textit{data table} lists the samples covered by the displayed rules.}
\label{fig:teaser}
\end{figure*}

In this section, we first introduce the dynamic construction of the rule hierarchy, and then present the design and implementation of the matrix-based hierarchical visualization that supports the exploration of rules. \looseness=-1

\subsection{Dynamic Hierarchy Construction}

Given the large-scale rule set extracted from a tree ensemble classifier, we need to organize the rules into a hierarchy to support exploration at different detail levels.
A simple way is to construct a static hierarchy with different numbers of rules extracted at each level to approximate the original model. 
However, such a practice lacks the flexibility to identify rules according to different analysis requirements. 
For example, in a stock price prediction model, most companies are predicted by the rules considering the historical price trend and volatility, while low-cap companies are more accurately predicted by the rules considering stock liquidity.
Therefore, it is desirable to build the rule hierarchy dynamically to better accommodate users' preference during exploration (\textbf{R2.2}). \looseness=-1

In particular, users start the exploration at the top level $L_0$.
The representative rules $R_0$ are displayed, which are extracted from all rules to approximate the model predictions on all samples, while the other rules are hidden.
Then, users can iteratively select rules of interest and zoom into their neighborhood for examination.
During this process, the rule hierarchy is dynamically constructed.
As illustrated in Fig.~\ref{fig:hierarchy}, the construction of level $L_{k+1}$ from level $L_k$ consists of three steps:\looseness=-1

\noindent
\textbf{Selecting a rule subset $R_k^{(sel)}$ from $R_k$}
and zooming in.
More representative rules should be displayed to illustrate the model behavior in the neighborhood of the selected rule subset $R_k^{(sel)}$.\looseness=-1

\noindent\textbf{Determining the neighborhood of $R_k^{(sel)}$} is the key in this process.
At the level $L_k$, the hidden rules are assigned to their nearest rules in $R_k$ based on the weighted Euclidean distance between the feature representations of rules.
Following the previous work~\cite{zhao2018iforest},
the weight of each attribute in the feature representation is calculated as the frequency of this attribute being used in the rules at this level, including representative and hidden ones.
The neighborhood of $R_k^{(sel)}$ consists of the rules in $R_k^{(sel)}$ and the hidden rules assigned to them.

\noindent\textbf{Extracting $R_{k+1}$} using model reduction to approximate the model predictions on the samples covered by the neighborhood of $R_k^{(sel)}$.
In order to keep the mental map between $L_k$ and $L_{k+1}$, we ensure that the user-selected rule subset $R_k^{(sel)}$ is displayed at $L_{k+1}$. 

\subsection{Visualization Design}

\begin{figure}[t]
  \centering
    \includegraphics[width=1\linewidth]{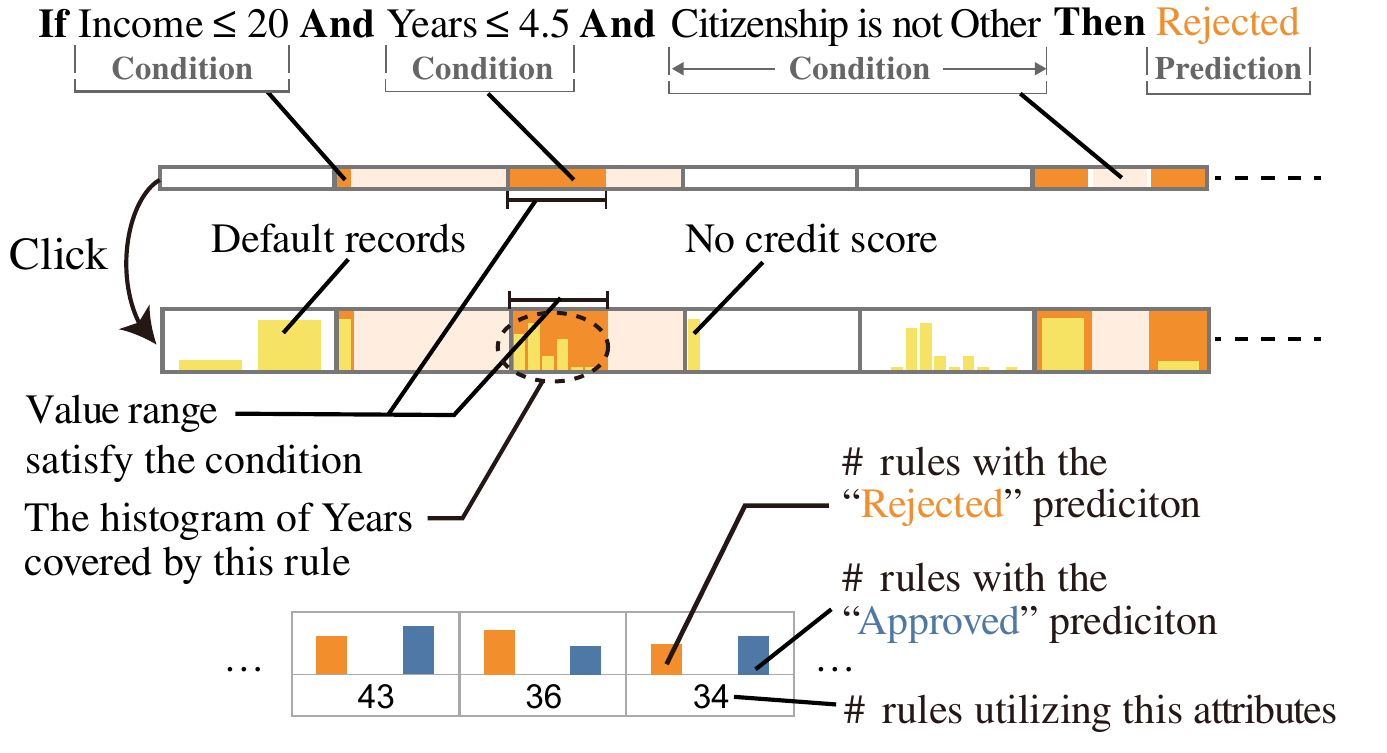}
    \put(-9, 125){(a)}
    \put(-9, 102){(b)}
    \put(-9, 73){(c)}
    \put(-9, 15){(d)}
    \vspace{-3mm}
    \caption{The design of the matrix visualization. \jwrevision{A rule (a) is visualized as a row (b) in the matrix, which can be expanded (c) to display sample distributions on different conditions. Additional rows (d) are appended at the bottom of the matrix to show cumulative statistics for each attribute.}}
    \label{fig:design}
\end{figure}
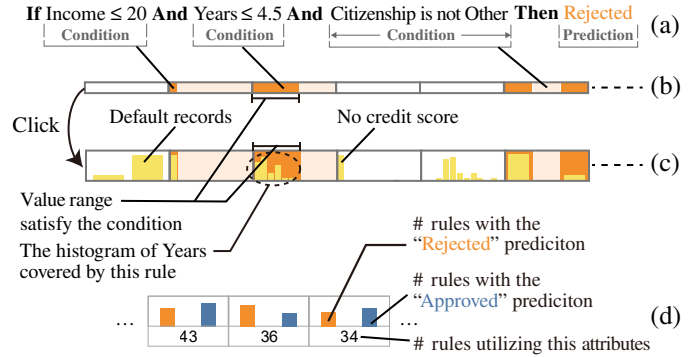

As shown in Fig.~\ref{fig:teaser}, the main component of {\system} is the \textit{matrix view} (Fig.~\ref{fig:teaser}(b)), \weikai{which uses an enhanced matrix visualization to provide an intuitive and informative overview of representative rules (\textbf{R2.1}).
We \jwrevision{adopt} 
the matrix visualization \jwrevision{because of} 
its effectiveness} in comparing and analyzing multiple rules.
\lizhen{\jwrevision{Moreover,} 
studies have shown that matrix visualizations outperform node-link diagrams and other techniques in terms of readability and scalability, particularly when dealing with large and complex datasets~\cite{alper2013weighted}.}
\weikai{However, existing methods focus only on numerical attributes, so we extend the matrix visualization} to accommodate categorical attributes.
As shown in Fig.~\ref{fig:design}(a), a rule is an \textbf{If-Then} statement that includes
1) a conjunction of conditions on attributes (\eg, $\text{Income} \le 20$ \textbf{AND} $\text{Years} \le 4.5$ \textbf{AND} \text{Citizenship is not Other}) 
and 2) a prediction label (\eg, \textcolor{orange}{Rejected}).
In the matrix visualization, each row represents a rule, and its color hue encodes the prediction label of this rule.
Each column represents an attribute.
If an attribute is used in a condition of this rule, the corresponding cell uses dark-shaded and light-shaded horizontal areas to indicate \weikai{the value ranges that satisfy and dissatisfy the condition, respectively} (Fig.~\ref{fig:design}(b)).
If an attribute is not used, the corresponding cell is left blank.
Such a design facilitates the comparison between the conditions of different rules.
The detailed attribute distributions of the samples covered by a rule are displayed when clicking on the corresponding row.
For example, in Fig.~\ref{fig:design}(c), the overlaid bar charts in each cell show that all the covered samples have no credit score, while most of them have default records.
This information helps users verify whether a rule utilizes the appropriate attributes to make its predictions.

We also supplement the matrix with additional rows and columns that provide auxiliary information for analysis.
At the bottom of the matrix, an additional row features cells that display the cumulative statistics for each attribute across the displayed rules.
As shown in Fig.~\ref{fig:design}(d), each cell in this row shows the number of rules utilizing this attribute, and includes a bar chart $\vcenter{\hbox{\includegraphics[height=1.5\fontcharht\font`\B]{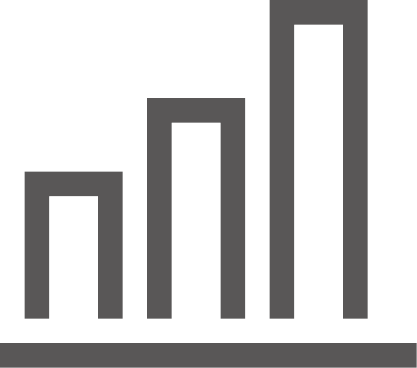}}}$ showing the prediction distribution of these rules. 
This helps users analyze the correlation of this attribute with different predictions.
On the left of the matrix, a glyph $\vcenter{\hbox{\includegraphics[height=0.5\fontcharht\font`\B]{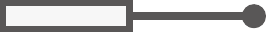}}}$ is placed next to each row.
The bar length represents the number of its neighborhood at this level.
When zooming in, the glyphs are also organized as an indented tree $\vcenter{\hbox{\includegraphics[height=1.5\fontcharht\font`\B]{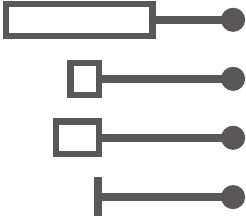}}}$ to show the hierarchical structure of the rules (Fig.~\ref{fig:case1-prior-default}D).
On the right of the matrix, there are three additional columns(Fig.~\ref{fig:teaser}H) that show the confidence, coverage, and anomaly score of each rule.
They represent the accuracy of rule prediction on its covered samples, the number of its covered samples, and the degree of its deviation from other rules, respectively.

In addition to the matrix view, {\system} provides three supporting views: \jwrevision{the \textit{attribute view} (Fig.~\ref{fig:teaser}(a)), the \textit{info view} (Fig.~\ref{fig:teaser}(c)), and the \textit{data table} (Fig.~\ref{fig:teaser}(d)).}
These views are coordinated with each other to facilitate interactive exploration. 

The \textit{attribute view} shows the attribute distributions of data samples. 
\lizhen{To enhance data exploration, we integrate scented widgets $\vcenter{\hbox{\includegraphics[height=1.5\fontcharht\font`\B]{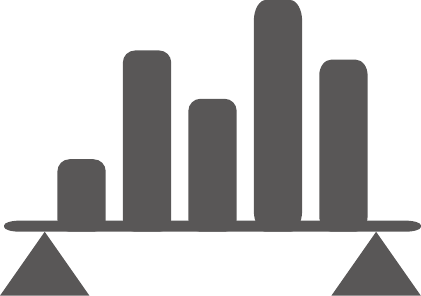}}}$~\cite{willett2007scented}, which provide visual cues \weikai{for easy} sample filtering (\textbf{R2.3}).}
\weikai{We chose scented widgets for their intuitiveness and ease of use, making them ideal for our users}.

The \textit{info view} presents the overall statistics of the representative rules and their covered samples at the current \jwrevision{hierarchical} level.
\weikai{This information reduces the analysis burden on understanding model behavior during exploration (\textbf{R2.2}).}
By selecting a rule in the \textit{matrix view} or applying filtering in the \textit{attribute view}, it also presents the details of the selected rule (Fig.~\ref{fig:teaser}G) or the class distribution change of the filtered samples (Fig.~\ref{fig:teaser}F), respectively.

The \textit{data table} lists the samples covered by the representative rules, \weikai{which directly presents the relationships between rules and training samples (\textbf{R2.3})}.
\weikai{The samples can also} be sorted based on their attributes to facilitate analysis.


\subsection{Interactive Exploration \shixia{and Analysis}}

We provide a set of interactions to facilitate 1) navigating through the rule hierarchy; 
\lizhen{2) analyzing rules of interest; and 3) analyzing attributes of interest } (\textbf{R2.3}).\looseness=-1

\subsubsection{Navigating through the rule hierarchy}
After identifying the rules of interest, users may want to compare them with their neighborhood for in-depth analysis.
This is supported by a dynamically constructed rule hierarchy, within which we implement the zoom function for navigating this hierarchy.
Specifically, users can click or brush the glyphs $\vcenter{\hbox{\includegraphics[height=0.5\fontcharht\font`\B]{figures/neighbor-glyph.pdf}}}$ of some rules. 
Then, {\system} dynamically extracts their child rules in their neighborhood.
Users can then zoom into the next level to examine these child rules.
In the \textit{matrix view}, the newly added rules are placed under their parent rules with their glyphs indented in comparison to those of their parent rules.
After zooming, the order of attributes may change since the number of rules utilizing each attribute differs between the two levels.
Some attributes may contribute more to predictions at this level.
We mark the attributes with the top-3 largest increases in their orders using upward arrows $\vcenter{\hbox{\includegraphics[height=\fontcharht\font`\B]{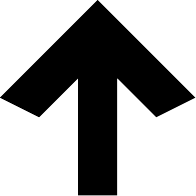}}}$ to inform users for further examination.
Moreover, by clicking the ``return'' button $\vcenter{\hbox{\includegraphics[height=\fontcharht\font`\B]{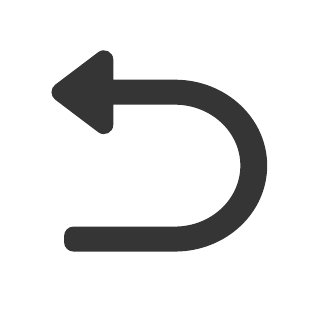}}}$ at the top-left corner of the \textit{matrix view}, users can navigate back to the previous level and analyze other rules of interest at that level.

\subsubsection{Analyzing rules of interest}
At each level of the hierarchy, users first get an overall understanding of the general model behavior, and then identify and analyze the rules of interest through interactive exploration.
By default, the rules are \jwrevision{sorted in descending order of their coverage} 
to highlight the common rules. 
They can also be sorted based on their confidence or anomaly scores by clicking on the corresponding column headers.
Moreover, to compare rules with similar conditions, users can \jwrevision{click on any attribute header to} \jwrevision{group} 
the rules based on the condition on \jwrevision{the attribute.}
For numerical attributes, the rules are sorted based on the lower and upper bounds of their \lizhen{value} ranges.
For categorical attributes, the rules with the same condition are grouped together.
Rules that do not utilize this attribute are placed at the bottom.



\begin{figure}[t]
  \centering
    \includegraphics[width=1\linewidth]{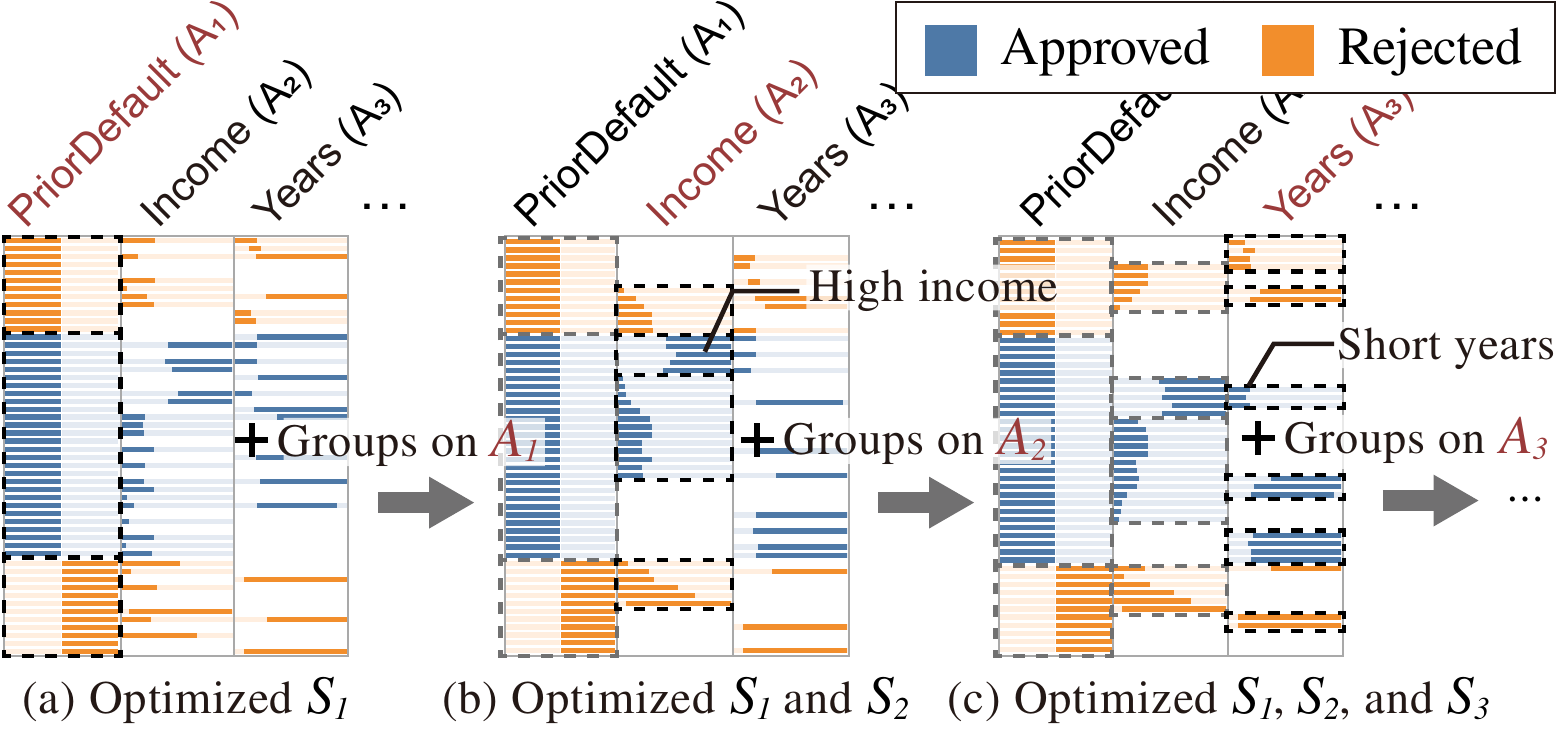}
    \vspace{-3mm}
    \caption{\jwrevision{An illustrative example of matrix reordering showing the formation of rule groups with (a) optimization on attribute $A_1$; (b) optimization on both $A_1$ and $A_2$; and (c) optimization on $A_1$, $A_2$ and $A_3$.}}
    \label{fig:changes}
\end{figure}

\weikai{In addition to grouping similar rules based on a single attribute, it is also desired to simultaneously consider multiple attributes.
For example, \jwrevision{compared to grouping rules only based on \emph{PriorDefault} (Fig.~\ref{fig:changes} (a))}, 
grouping rules with similar conditions on \emph{PriorDefault} and \emph{Income} together (Fig.~\ref{fig:changes} (b)) \jwrevision{makes} it more readily to discover the pattern that \textit{``applicants with no prior default and high-income tend to be accepted"}. 
Therefore, \jwrevision{to facilitate the discovery of rule patterns,} we develop a matrix reordering algorithm \jwrevision{to group rules based on multiple attributes.} 
}


\lizhen{The basic idea 
\weikai{
is to group rules with consistent predictions and similar conditions \jwrevision{together}.}
\weikai{To achieve that, we maximize the number of similar conditions on the specific attributes between adjacent rules.}
} 
\jwrevision{Specifically, given $n$ rules $R = [r_1, r_2, ..., r_n]$, and $m$ attributes $A_1$, $A_2$, ..., $A_m$; we maximize $S_1$, $S_2$, ..., $S_m$, where $S_j$ represents the number of similar conditions on attribute $A_j$ between adjacent rules:}

\begin{equation}
S_j = \sum_{i=1}^{n-1} \mathbb{I}(\text{pred}_{\pi(i)}=\text{pred}_{\pi(i+1)}) * \text{sim}_{j}(r_{\pi(i)}, r_{\pi(i+1)}).
\end{equation}

\noindent Here, \jwrevision{$\pi(i)$ is the index of the rule in $R$ which is reordered to row $i$ in the matrix.
\weikai{The first term} is an indicator function, which equals $1$ when the predictions of the two rules are consistent, \weikai{and $0$ otherwise}.}
\weikai{The second term}
$\text{sim}_{j}(r_{\pi(i)}, r_{\pi(i+1)})$ is also an indicator function that determines whether the \jwrevision{two rules have similar conditions on the $j$th attribute.} 
\jwrevision{For categorical attributes,} it is defined as whether the conditions are the same.
\jwrevision{For} numerical attributes, it is defined as whether \jwrevision{both} the left and right endpoints \jwrevision{between the two conditions} 
differ \jwrevision{within a user-specified threshold $\tau$.} 
\weikai{We set $\tau=0.1$ by default.}
\lizhenrev{This threshold controls whether two numerical conditions are similar enough to be placed adjacently. A smaller $\tau$ results in stricter matching criteria, while a larger $\tau$ value relaxes the criteria.}

Maximizing $S_j$ is equivalent to solving a traveling salesman problem\lizhenrev{~\cite{applegate2006traveling}}, which will reorder the rules and form rule groups on the attribute $A_j$.
Fig.~\ref{fig:changes} shows examples of rule groups formed on each attribute.
\jwrevision{\weikai{However, reordering rules to optimize all $S_j$ simultaneously is impossible as different optimization goals may conflict with each other.} 
As experts are more concerned with patterns on important/preferred attributes, we adopt the idea of \lizhen{the constraint method}~\cite{deb2016multi}, which maximizes $S_1, S_2, \ldots, S_m$ in turn. 
\weikai{The constraints are that when maximizing $S_j$, the reordering of rules in a formed rule group can only be carried out within the group without crossing the group boundaries.} 
This ensures minimal changes to $S_1, S_2, \ldots, S_{j-1}$ when maximizing $S_j$.} 
\weikai{We use the divide-and-conquer strategy proposed by Liu~\etal\cite{liu2017towards} to solve this constrained traveling salesman problem.}

\subsubsection{Analyzing attributes of interest}
As the number of attributes may exceed several hundred, it is impractical to directly display all attributes.
To address this issue, we sort the attributes based on their importance and organize them into pages.
On the front page, the most frequently used attributes are presented, which are important to the model prediction.
Users can navigate to subsequent pages for attributes that are less frequently used according to 
their analysis needs.
Moreover, users can pin specific attributes to the front page and adjust their positions by drag-and-drop.
This facilitates the comparison of multiple attributes and understanding of their collaborative behavior.
\weikai{For example, Fig.~\ref{fig:changes}(c) shows that by placing attributes \emph{Income} and \emph{Years} adjacently, users can identify patterns such as high income often leading to credit card approvals despite short working years.}
\section{Evaluation}
\label{sec:evaluation}

\begin{table*}[t]
\setlength{\tabcolsep}{8.5pt}
\vspace{-0.5mm}
\caption{Performance comparison between our method and the baseline methods in terms of fidelity and average anomaly score.} 
\resizebox{\linewidth}{!}{
\begin{tabular}{cccccccccccccc} 
\toprule 
\multirow{2}{*}{Dataset} & \multirow{2}{*}{\#Classes} & \multirow{2}{*}{\lizhenrev{\#Attributes}} & \multirow{2}{*}{\#Samples} & \multirow{2}{*}{Model} & \multirow{2}{*}{\lizhenrev{\#Rules}} & \multicolumn{2}{c}{Ours} & \multicolumn{2}{c}{RuleMatrix} & \multicolumn{2}{c}{HSR} & \multicolumn{2}{c}{Node harvest} \\ 
\cmidrule(lr){7-8} \cmidrule(lr){9-10} \cmidrule(lr){11-12} \cmidrule(lr){13-14}
~ & ~ & ~ & ~ & ~ & ~ & Fidelity & AS & Fidelity & AS & Fidelity & AS & Fidelity & AS \\
\midrule
\multirow{2}*{Credit Card} & \multirow{2}*{2} & \multirow{2}*{\lizhenrev{14}} & \multirow{2}*{690} & RF & \lizhenrev{7,085} & \textbf{0.9538} & \textbf{0.1309} & 0.9503 & 0.0880 & 0.9264 & 0.0000 & 0.9300 & 0.0473\\
~ & ~ & ~ & ~ & GBT & \lizhenrev{2,126} & \textbf{0.9249} & \textbf{0.1271} & 0.9075 & 0.0735 & 0.8682 & 0.0846 & 0.9017 & 0.0042\\
\hline
\multirow{2}*{Wine Quality} & \multirow{2}*{2} & \multirow{2}*{\lizhenrev{11}} & \multirow{2}*{1,599} & RF & \lizhenrev{28,116} & \textbf{0.8725} & 0.1432 & 0.8665 & 0.0689 & 0.7484 & 0.0000 & 0.5175 & \textbf{0.1660}\\
~ & ~ & ~ & ~ & GBT & \lizhenrev{7,474} & \textbf{0.8350} & \textbf{0.0974} & 0.8065 & 0.0329 & 0.7368 & 0.0000 & 0.8275 & 0.0299\\
\hline
\multirow{2}*{Crime} & \multirow{2}*{2} & \multirow{2}*{\lizhenrev{127}} & \multirow{2}*{1,994} & RF & \lizhenrev{12,576} & \textbf{0.9178} & \textbf{0.1999} & 0.9042 & 0.0203 & 0.8910 & 0.0000 & 0.8577 & 0.1602\\
~ & ~ & ~ & ~ & GBT & \lizhenrev{4,322} & \textbf{0.9118} & \textbf{0.1152} & 0.8986 & 0.0119 & 0.8368 & 0.0000 & 0.8978 & 0.0489\\
\hline
\multirow{2}*{Abalone} & \multirow{2}*{4} & \multirow{2}*{\lizhenrev{8}} & \multirow{2}*{4,177} & RF & \lizhenrev{23,500} & \textbf{0.8392} & \textbf{0.1247} & 0.8243 & 0.0085 & 0.6695 & 0.0081 & N/A & N/A\\
~ & ~ & ~ & ~ & GBT & \lizhenrev{11,810} & \textbf{0.8967} & 0.1263 & 0.8268 & 0.1231 & 0.5715 & \textbf{0.3553} & N/A & N/A\\
\hline
\multirow{2}*{Obesity} & \multirow{2}*{7} & \multirow{2}*{\lizhenrev{16}} & \multirow{2}*{2,111} & RF & \lizhenrev{19,490} & \textbf{0.8561} & \textbf{0.1732} & 0.7720 & 0.0138 & 0.7277 & 0.0683 & N/A & N/A\\
~ & ~ & ~ & ~ & GBT & \lizhenrev{30,458} & \textbf{0.9470} & \textbf{0.2309} & 0.8394 & 0.0004 & 0.7233 & 0.0084 & N/A & N/A\\
\hline
\multirow{2}*{Dry Bean} & \multirow{2}*{7} & \multirow{2}*{\lizhenrev{16}} & \multirow{2}*{13,611} & RF & \lizhenrev{6,209} & \textbf{0.9738} & \textbf{0.2829} & 0.9702 & 0.0032 & 0.8726 & 0.0193 & N/A & N/A\\
~ & ~ & ~ & ~ & GBT & \lizhenrev{52,455} & \textbf{0.9151} & \textbf{0.3032} & 0.9151 & 0.0946 & 0.8369 & 0.2360 & N/A & N/A\\
\bottomrule
\end{tabular}
}
\begin{tablenotes}
\item {RF}: random forest. {GBT}: gradient boosted tree.
{AS}: average anomaly score. Node harvest cannot handle multi-classification datasets.
\end{tablenotes}
\label{tab:result}
\end{table*}

We \jwrevision{first} conducted a quantitative experiment to evaluate the effectiveness of the anomaly-biased model reduction method. 
\jwrevision{We} then presented two case studies to showcase how {\system} helps domain experts understand the decision logic of tree ensemble classifiers and improve them.
\looseness=-1

\subsection{Quantitative Experiment}

\noindent\textbf{Datasets}.
We conducted the quantitative experiment on six datasets from the UCI Machine Learning Repository~\cite{asuncion2007uci}, as used in previous works~\cite{ming2018rulematrix, yuan2024visual}, including three binary classification datasets (Credit Card, Wine Quality, Crime) and three multiple classification ones (Abalone, Obesity, Dry Bean). \lizhenrev{The attribute numbers across these datasets range from 8 to 127, while the sample numbers range from 690 to 13,611. The diverse characteristics of these datasets enable a robust evaluation across various tasks and dataset scales.}\looseness=-1

\noindent\textbf{Baseline methods}.
We selected three baseline methods, including
two model surrogate methods, RuleMatrix~\cite{ming2018rulematrix} and HSR~\cite{yuan2024visual}, and a state-of-the-art model reduction method, node harvest~\cite{meinshausen2010node}.
RuleMatrix generates a sequential rule list, while HSR generates a hierarchical rule set.
Different from these two methods, node harvest directly extracts the rules from the original model rather than generating new ones.
However, it cannot be used in multi-class classification.

\noindent\textbf{Measures}.
We evaluated the quality of the methods from two aspects.
First, \emph{fidelity} measures the ability to preserve the behavior and predictive accuracy of the original model.
It is defined as the ratio of the samples with the same predictions produced by the generated rule sets and the original model. 
Second, the \emph{average anomaly score} evaluates the ability to preserve anomalous rules.

\noindent\textbf{Experiment settings}.
To demonstrate the generalizability of our method to different tree ensemble classifiers, we conducted experiments
on random forest models and gradient boosted tree models.
Each dataset was partitioned into a 75\% training set and a 25\% test set.
We first trained each model on the training set, 
\lizhen{ resulting in models with rule counts ranging from 2,126 (LightGBM on credit card dataset) to 52,455 (LightGBM on drybean dataset). }
\lizhenrev{This wide range of rule numbers ensures a thorough evaluation across different levels of model complexity}.
Then, we performed model reduction to approximate the predictions on the training set.
\lizhen{ All methods are required to select 80 rules (consistent with Sec. IV) for this approximation. }
The fidelity was calculated on the test set, and the average anomaly score was calculated on the generated rule set.
We ran five trials for each setting to reduce the effect of randomness.


\noindent\textbf{Results}.
Our results are summarized in Table~\ref{tab:result}.
Our method achieved the highest fidelity across all cases,
and also achieved higher average anomaly scores than the baseline methods except in two cases. 
On the Wine Quality dataset, node harvest obtained a higher average anomaly score for the random forest model,
while on the Abalone dataset, HSR obtained a higher average anomaly score for the gradient boosted model.
However, the fidelity scores achieved in these two cases are significantly lower ($0.5175$ and $0.5715$), indicating the corresponding rule sets failed to approximate the behavior of the original model.
In contrast, our method achieved the highest fidelity and the second-highest average anomaly scores. 
This demonstrates that our method can strike a balance between fidelity and the preservation of anomalous rules.

\subsection{Case Studies}
We conducted two case studies to show the usefulness of {\system} in explaining and diagnosing tree ensemble classifiers.\looseness=-1

\subsubsection{Credit Card Approval}

In this case study, we collaborated with E$_1$ and E$_2$ to illustrate how {\system} improves the understanding and diagnosis of the model to approve credit card applications.
E$_1$ is a bank account manager with two years of work experience, who aims to increase credit card approval rates while minimizing the risk of defaults.
E$_2$ is a data scientist responsible for developing the credit card approval model and improving its performance.
\weikai{\jwrevision{Both E$_1$ and E$_2$}
have participated in the interview for collecting design goals in Sec.~\ref{sec:overview}}.
\lizhenrev{During the study, E$_1$ interacted with the system to explore and identify potential model issues, while E$_2$ helped explain the underlying causes of these findings from a machine learning perspective and suggested ways to improve the model.}
The study used the Australian Credit Approval dataset~\cite{australian_credit_approval}, which consists of 690 samples, each representing a credit card applicant with 16 attributes and a label indicating approval or rejection.
The dataset includes 307 ``\textbf{approved}'' samples and 383 ``\textbf{rejected}'' samples.
These samples were randomly divided into training and test data with a ratio of 75\% and 25\%, respectively.
To comprehensively evaluate the model, E$_2$ used Synthetic Data Vault~\cite{SDV} to augment the test data based on its distribution and labeled the new test samples in collaboration with E$_1$.
\lizhenrev{The detailed augmentation process can be found in the supplemental material.}
After the augmentation, there are 500 samples in the test data.
A random forest model \lizhen{with 200 trees and a maximum depth of 10} has been trained on this dataset with the scikit-learn library~\cite{scikit-learn}.
The trained model includes \textbf{7,085} rules and achieves an accuracy of 83.60\%.

\noindent\textbf{Overview}.
Initially, E$_1$ investigated the attributes frequently used in the extracted rules to see whether the decision-making process was reasonable.
As shown in Fig.~\ref{fig:teaser}A, the attributes that describe income and credit status, such as \emph{PriorDefault}, \emph{Job}, \emph{Income}, \emph{Years Employed}, are frequently used and thus appear in the first few columns.
This observation aligns with his experience, as these attributes are more important in credit card approval than those less frequently used 
attributes like \emph{Gender} and \emph{DriverLicenses}.
Subsequently, E$_1$ examined the rules to see how these frequently used attributes influenced model predictions.
He focused on two types of rules: common \jwrevision{rules} and anomalous rules.

\begin{figure}[t]
\centering
\includegraphics[width=1\linewidth]{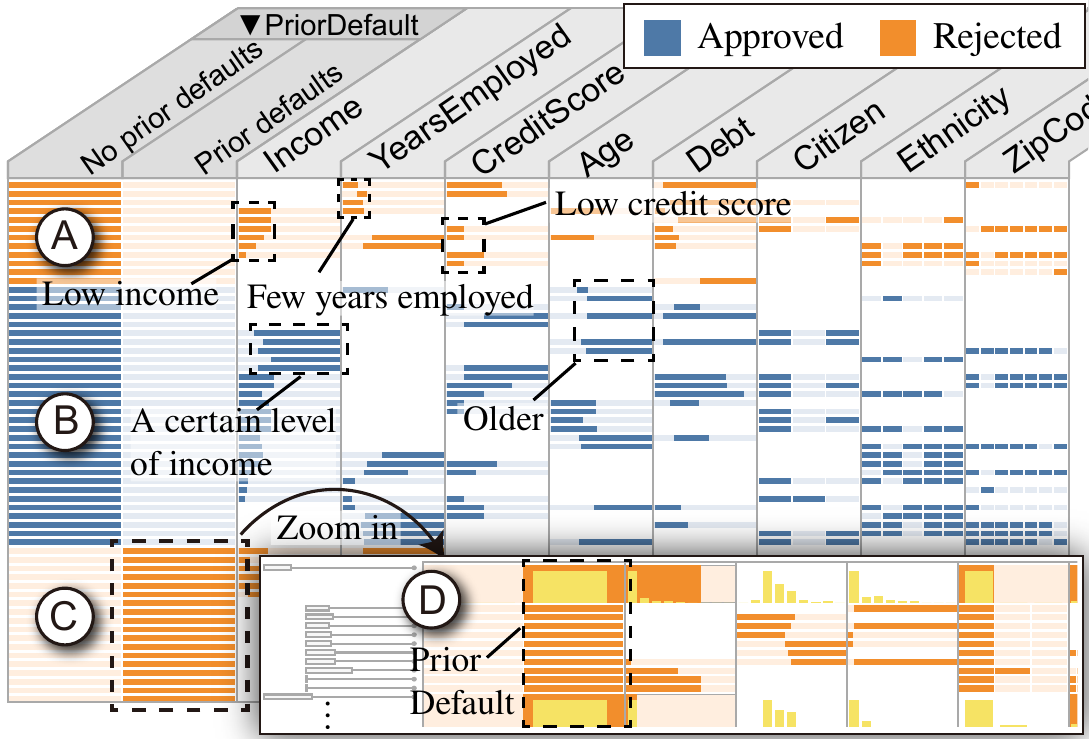}
\caption{Anayzing the rules that utilize the attribute \emph{PriorDefault}:
\lizhen{rules with no prior defaults may reject (A) or approve (B) credit card applications, while those with prior defaults consistently reject applications (C). Zoom in region C to examine the neighborhood (D). }
} 
\label{fig:case1-prior-default}
\end{figure}

\noindent\textbf{Analyzing common rules}.
By default, the rules are sorted in descending order of their coverage, \jwrev{which facilitates the analysis of common rules (rules with higher coverage).}
\jwrev{E$_1$} found that most of \jwrev{the common rules} 
used the attribute \emph{PriorDefault} (Fig.~\ref{fig:teaser}B).
He wanted to examine how \jwrev{this attribute} 
influenced the model predictions. 
To this end, he clicked on the header of \emph{PriorDefault}, and \jwrev{the matrix reordering algorithm grouped the rules based on their predictions and their conditions on this and subsequent attributes}. 
There are three groups 
(Fig.~\ref{fig:case1-prior-default}). 
\jwrev{Rules with the condition of no prior defaults but predictions of ``rejected'' (Fig.~\ref{fig:case1-prior-default}A) linked to negative conditions such as low income, few years employed, or low credit scores. Rules with the condition of no prior defaults and predictions of 
``approved'' (Fig.~\ref{fig:case1-prior-default}B) had additional positive conditions, such as a certain income level or being older, indicating a lower risk of default and a more stable status.} 
E$_1$ regarded the behavior (Fig.~\ref{fig:case1-prior-default}A and B) as reasonable, since the condition of no prior defaults cannot directly lead to the approval, and it should be combined with additional conditions for further consideration.
\jwrev{The third group included rules with the condition of prior defaults and predictions of ``rejected'' (Fig.~\ref{fig:case1-prior-default}C).
Since there are some real-world cases where applicants with prior default records are still approved, E$_1$ wanted to see whether these were reflected.} 
\jwrev{He selected the rules in (Fig.~\ref{fig:case1-prior-default}C) and zoomed into their neighborhood in the dynamic hierarchy, and found that all rules in the neighborhood had predictions of ``rejected'' (Fig.~\ref{fig:case1-prior-default}D).}
\jwrev{Consultation with} E$_2$ 
\jwrev{clarified} that a credit card approval model typically rejects all applicants with prior default records in realistic scenarios and leaves disputed cases for manual processing.
To summarize, E$_1$ was satisfied with the behavior of the rules utilizing \emph{PriorDefault}.
Similarly, E$_1$ analyzed other frequently used attributes, including \emph{Job}, \emph{Income}, \emph{Years Employed}, and acknowledged \jwrev{their proper behavior in making predictions.} 

E$_1$ continued to examine common rules and noticed a rule that had significantly lower confidence than the rules with similar coverage (Fig.~\ref{fig:teaser}C).
The low confidence indicated that this rule covered many mispredicted samples.
\jwrev{The \emph{data table} enabled him to examine the samples covered by this rule.}
\jwrevision{He} found that these samples shared some negative conditions (Fig.~\ref{fig:teaser}D), including unemployment, missing credit scores, lack of income, and having prior default records, \jwrev{but were labeled as ``approved'' (Fig.~\ref{fig:teaser}E).} 
E$_1$ suspected that they 
should be rejected given those negative conditions.
He then \jwrev{used the \emph{attribute view}} to filter the samples with such negative conditions and found \jwrev{in the \emph{info view}} that they were mostly labeled as ``rejected.''
This indicated that there were inconsistent approval criteria in the training data.
He raised this issue with E$_2$, and E$_2$ commented that such inconsistency harmed the model training and led to performance degradation.
Therefore, E$_1$ unified the approval criteria in these samples and corrected their labels to be ``rejected.''
By retraining the model on the corrected data, the accuracy increased from 83.60\% to 86.20\%, and the prediction confidence increased from 0.7403 to 0.7423.

\begin{figure}[t]
\centering
\includegraphics[width=1\linewidth]{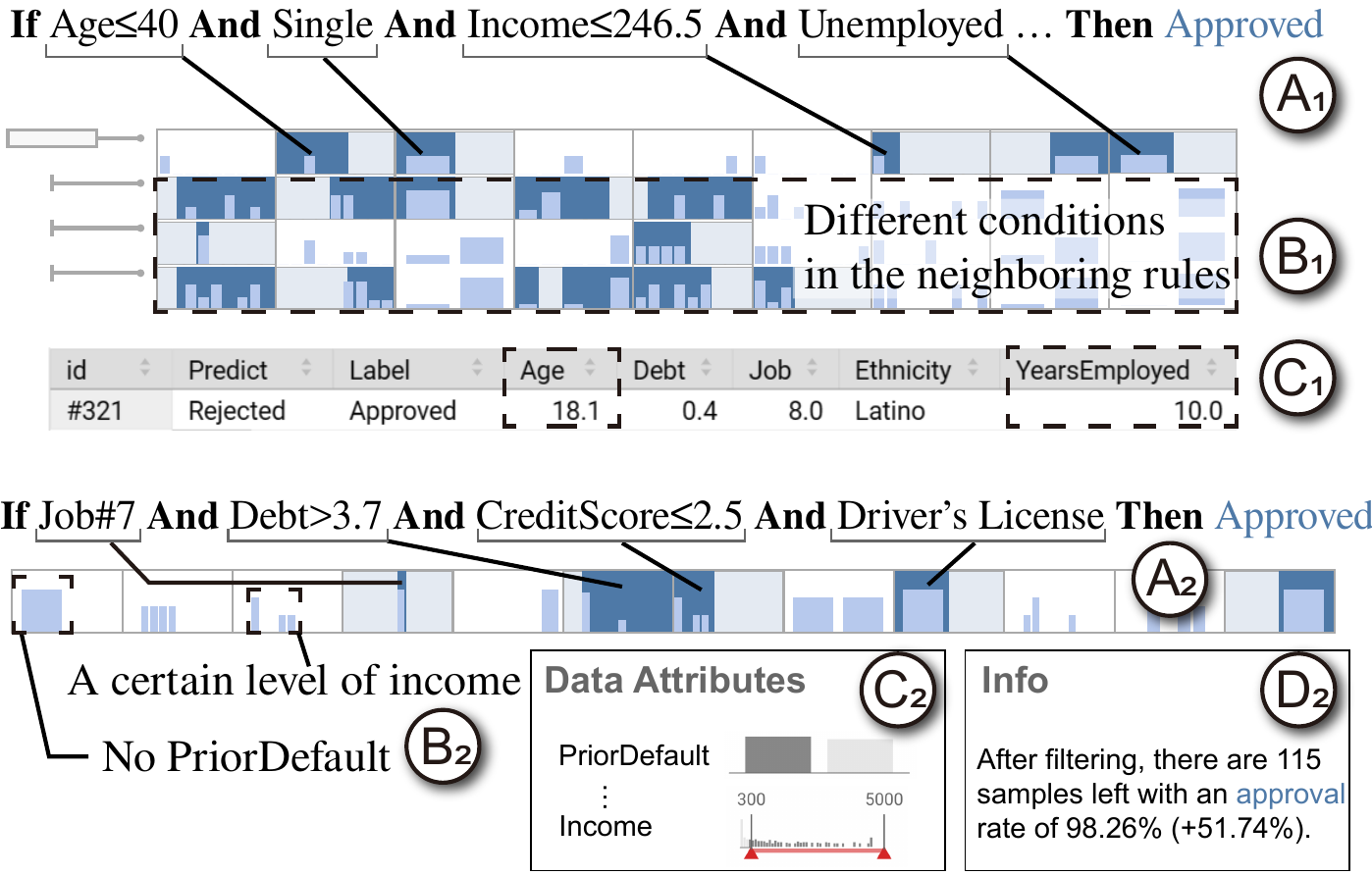}
\caption{Analyzing the top-2 rules with the highest anomaly scores: 1)
\lizhen{the analysis of the first rule (A$_1$), its neighbors (B$_1$), and the associated sample (C$_1$) reveals a labeling error; 2) the analysis of the second rule (A$_2$), its sample distribution (B$_2$), and information from the \textit{attribute \jwrevision{and info} views} (C$_2$, D$_2$) confirms a spurious association.}}
\vspace{-4mm}
\label{fig:case1-anomaly}
\end{figure}

\noindent\textbf{Analyzing anomalous rules}.
\jwrev{Reordering rules according to their anomaly scores facilitates the analysis of anomalous rules.} 
E$_1$ started his analysis by examining the rule with the highest anomaly score.
As shown in Fig.~\ref{fig:case1-anomaly}A$_1$, this rule suggests approval for young, single, unemployed applicants with a medium-to-low income.
These conditions were mostly neutral or negative, which should not lead to approval.
He zoomed into 
its neighborhood and observed that \jwrev{few conditions} were shared 
(Fig.~\ref{fig:case1-anomaly}B$_1$), which also indicated that its behavior differed a lot from other rules.
\jwrev{He then examined} its covered samples \jwrev{in the \emph{data table}}. As shown in Fig.~\ref{fig:case1-anomaly}C$_1$, it covered only an 18-year-old applicant with a 10-year work experience (\#321), \jwrev{which was suspicious}.
E$_1$ confirmed that this applicant should not be approved \jwrevision{and corrected its label to ``rejected''.} 

Subsequently, E$_1$ examined the rule with the second-highest anomaly score.
This rule suggested approval for applicants with \jwrev{primarily negative or neutral conditions such as some debt, low credit score, having a driver's license (Fig.~\ref{fig:case1-anomaly}A$_2$). 
E$_1$ \jwrevision{doubted the approval decision for these applicants, and checked the samples covered by this rule. \jwrev{He found} most of the applicants had no prior default records and a certain level of income (Fig.~\ref{fig:case1-anomaly}B$_2$).
}}
E$_1$ believed 
these positive conditions were the real reason for the approval.
By filtering samples with such positive conditions using the scented widgets in the \textit{attribute view} (Fig.~\ref{fig:case1-anomaly}C$_2$), he found that most of them (\lizhenrev{98.26\%}) are approved (Fig.~\ref{fig:case1-anomaly}D$_2$).
Thus, he \jwrevision{suspected} \jwrev{the occurrence of this rule is}
because of the spurious association between the irrelevant conditions and the approval decision in these covered samples (Fig.~\ref{fig:case1-anomaly}A$_2$). 
\jwrev{Discussion with E$_2$ suggested}
adding more diverse samples, especially 
\jwrevision{those} that meet the conditions of this rule but get rejected, 
to mitigate this spurious association.
\jwrev{E$_2$ then} generated \jwrevision{more} 
samples using data augmentation, and \jwrev{E$_1$} validated these samples to ensure they accurately reflected real-world scenarios. 

They analyzed other anomalous rules with the top-20 anomaly scores and found 1 more rule caused by label error and 15 more rules caused by spurious associations.
Similarly, they corrected the label error and augmented the training samples.
After retraining the model, the accuracy increased from 86.20\% to 88.60\%, and the average prediction confidence increased from 0.7423 to 0.7630.
The average anomaly score of the rules also decreased from 0.0547 to 0.0493.

\begin{figure*}[t]
  \centering
    \includegraphics[width=1\linewidth]{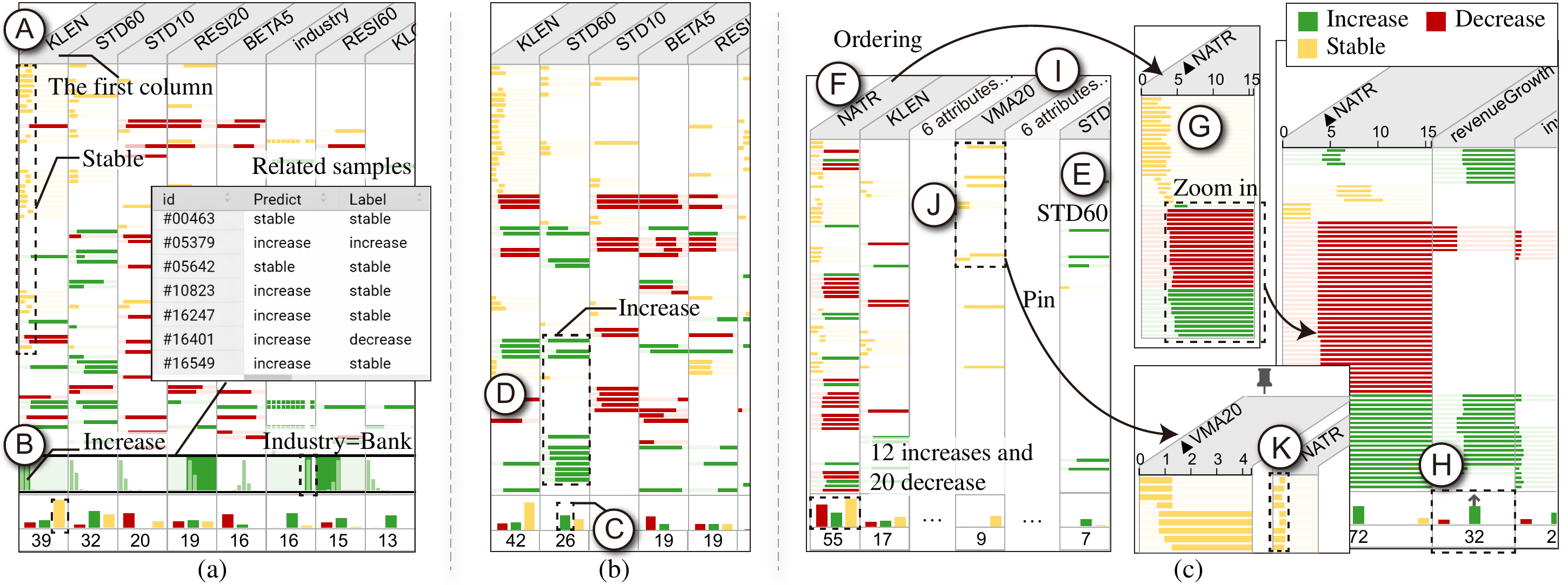}
     \vspace{-2mm}
        \caption{
        \lizhen{
        Analyzing rules for quantitative stock trading: (a) in the initial model, rules with lower \textit{KLEN} typically indicate ``Stable'' (A), except for one outlier (B); (b) after removing attributes that cause biased predictions, \textit{STD60} was found to be crucial for predicting an ``Increase'' (C, D); (c) after adding more useful attributes for predicting ``Increase'', the importance of \textit{STD60} decreased (E), while \textit{NATR} became the most important attribute (F), which triggered further analysis (G, H, I).
        }}
    \vspace{-3mm}
    \label{fig:case2-stock-step1}
\end{figure*}

\subsubsection{Quantitative Stock Trading}
In this case study, we collaborated with E$_3$, a junior quantitative stock trader specializing in short-term trading in the U.S. stock market.
\weikai{He has also participated in the interview in Sec.~\ref{sec:overview}}.
His goal was to identify stocks with rising prices using a stock price prediction model.
The training data consists of 119,416 samples collected from the U.S. stock market from April 2020 to April 2023, and the test data consists of 10,062 samples collected from May 2023 to August 2023. 
Each sample represents the information of a stock at a specific time point, including two types of attributes~\cite{edwards2018technical}: 158 \emph{technical analysis attributes} and 99 \emph{fundamental attributes}.
The technical analysis attributes are derived from historical prices to evaluate and forecast financial market trends.
For example, \emph{STD60} measures the variance in stock prices over the preceding 60 days.
The fundamental attributes provide insights into a company's financial health and operational performance.
For example, \emph{revenueGrowth} reflects the company's revenue growth rate.
Based on the stock price fluctuation over the next seven days, the samples are categorized  into three classes: ``\textbf{increase},'' ``\textbf{decrease},'' and ``\textbf{stable}.'' 
Directly using all attributes to train the model can deteriorate its generalization ability due to attribute redundancy~\cite{htun2023survey}. 
Following the common practice, E$_3$ \jwrevision{selected attributes with an information coefficient larger than 0.05~\cite{goodwin1998information}, resulting in 140 attributes remaining.} 
He then used these attributes to train a gradient boosted tree model with LightGBM~\cite{ke2017lightgbm} due to its superior performance and widespread adoption. 
\lizhen{The model was configured with 1000 trees and a maximum depth of 14.}
This model includes \textbf{81,000} rules but achieves an accuracy of only 56.26\%, with class-specific accuracy of 26.69\% for ``increase,'' 41.53\% for ``decrease,'' and 66.56\% for ``stable.''

\noindent\textbf{Overview}. 
Initially, E$_3$ wanted to verify whether the model utilized appropriate attributes for prediction. 
He checked the most frequently used attributes in the \textit{matrix view} and 
found the top-5 important attributes are technical analysis attributes (Fig.~\ref{fig:case2-stock-step1}(a)).
\jwrev{E$_3$ considered it reasonable as technical analysis attributes offered a more direct reflection of price fluctuations than fundamental attributes. 
Reordering rules based on the most important attribute \emph{KLEN}, he found that the model tended to predict the stock price as ``stable'' when \emph{KLEN} was low. This is reasonable because stocks with lower \emph{KLEN} are more likely to maintain stable prices in the next few days.}

\noindent\textbf{Addressing biased predictions}.
\jwrev{However, there was an exception: when \emph{KLEN} was low, one rule made an ``increase'' prediction (Fig.~\ref{fig:case2-stock-step1}B).}
\jwrevision{Examining} the samples covered by this rule, 
\jwrev{E$_3$} found that these samples had small \emph{KLEN}, medium \emph{RESI20}, and medium \emph{RESI60}, all of which indicated stable stock prices.
\jwrevision{However, \jwrev{they also shared} the condition of \emph{industry=Bank}, \jwrev{which} may be the reason for the prediction of ``increase.''} 
\jwrev{E$_3$ explained that 
the banks associated with these samples experienced a growth period when the training data was collected, so the trained model tended to predict them as ``increase.''}
However, such a pattern did not exist in the test samples and thus caused a performance drop.
To address this issue, he removed the attribute \jwrevision{\emph{industry}.} 
\jwrevision{After retraining the model,} 
he observed an improvement in accuracy from 56.26\% to 56.52\%.
Specifically, on the samples with \emph{industry=Bank}, the accuracy increased from 48.90\% to 51.22\%.

\noindent\textbf{Improving model performance in ``increase.''}
As E$_3$'s primary goal was to find stocks with possibly rising prices to maximize profits, he focused on analyzing the attributes \lizhenrev{associated} to the ``increase'' \lizhenrev{prediction} in the retrained model. 
\jwrevision{As shown in Fig.~\ref{fig:case2-stock-step1}C, he observed that \emph{STD60} was frequently used for the prediction of ``increase''.}
\jwrev{Reordering rules based on this attribute,} 
he noticed that the rules with the conditions of larger \emph{STD60} consistently made predictions of ``increase'' (Fig.~\ref{fig:case2-stock-step1}D).
However, \jwrev{as E$_3$ commented}, a larger \emph{STD60} just meant a large fluctuation in stock prices over the past 60 days, which did not necessarily indicate an increasing stock price.
\jwrev{E$_3$ considered \emph{STD60} was not a useful attribute for the ``increase'' prediction, and suspected the heavy reliance on it accounted for the low prediction accuracy in this class (26.35\%).}
\lizhen{\jwrevision{He then} 
performed feature engineering by leveraging TA-lib~\cite{ta-lib}, an open-source tool, to generate \weikai{more} technical analysis attributes.}
After adding these attributes and retraining the model, the overall accuracy increased from 56.52\% to 58.35\%, and the class accuracy of ``increase'' increased from 26.35\% to 28.24\%.


To verify whether the newly added attributes were used correctly, E$_3$ examined \jwrev{the cumulative statistics for attributes at the bottom of the \emph{matrix view}.} 
\weikai{The usage of \emph{STD60} decreased from 26 to 7 (Fig.~\ref{fig:case2-stock-step1}E), while a newly added attribute, \emph{NATR}, became the most frequently used (Fig.~\ref{fig:case2-stock-step1}F).
\jwrev{\emph{NATR}} represents asset volatility.}
\jwrevision{A high \emph{NATR} indicates a large variation in stock prices. Reordering rules based on \emph{NATR}, E$_3$ found that, indeed, high \emph{NATR} values are associated with ``increase'' or ``decrease'' predictions, while low \emph{NATR} values are associated with ``stable'' predictions (Fig.~\ref{fig:case2-stock-step1}G).} 
\jwrev{Examining the neighborhood of rules with high \emph{NATR},}
\jwrev{E$_3$} found that \emph{revenueGrowth}, \jwrevision{the growth of the profit rate,} 
was frequently used \jwrev{together} 
(Fig.~\ref{fig:case2-stock-step1}H).
\weikai{\jwrev{Reodering rules based on \emph{NATR} and \emph{revenueGrowth}} 
revealed a clear pattern that} high \emph{revenueGrowth} mostly led to an ``increase'' prediction, while low \emph{revenueGrowth} mostly led to a ``decrease'' prediction.
This observation was consistent with his experience.

\noindent\textbf{Removing redundant attributes}. 
During further exploration, E$_3$ identified some redundant attributes for prediction.
For example, \emph{VMA20}, \lizhen{which indicates the average trading volume in the preceding 20 days}, \jwrev{was recognized as an important attribute 
(Fig.~\ref{fig:case2-stock-step1}I).}
However, \jwrev{rules using this attribute made predictions of ``stable'' 
no matter low or high \emph{VMA20} values,} 
\jwrevision{which seemed to suggest the value of VMA20 does not affect the prediction}. 
\jwrev{Pinning \emph{VMA20} to the first column and \jwrev{reordering} rules based on it,}
\jwrev{E$_3$} found that \jwrevision{all the rules using VMA20 also \jwrev{
had low \emph{NATR} values.} As low NATR indicates a small variation in stock prices, E$_3$ believed it was the actual reason for the predictions of these rules, and \emph{VMA20} was redundant. He proposed to remove \emph{VMA20}.} 
Similarly, he removed another eight redundant attributes.
After retraining the model, its accuracy was improved to 59.20\%.

Finally, E$_3$ employed the final model for simulated trading.
The trading strategy involved selecting 20 stocks predicted to be the most likely to increase in each 7-day trading cycle over a period of 3 months.
Backtesting proved that this trading strategy produced an excess return of 48.7\% with a maximum drawdown of 5.0\%.
E$_3$ was pleased with this performance and said the comprehensive understanding of this model made him more confident in employing \jwrevision{the model} 
for real-world trading.

\subsection{Expert Feedback and Discussion}
After the case studies, we conducted semi-structured interviews with each of the six experts \lizhen{(E$_1$-E$_6$) introduced in Sec.~\ref{sec:overview}-A.}
\lizhenrev{E$_4$-E$_6$ were not involved in the case studies. These newly involved experts included a fund manager with two years of experience (E$_4$), a vice president at a commercial bank branch (E$_5$), and a machine learning researcher with five years of experience (E$_6$). For E$_4$-E$_6$, we spent the initial 20 minutes introducing {\system} and presenting the case studies. Each expert then used {\system} for model understanding and verification tasks. It took 5-15 minutes for them to become familiar with the interactions and conduct analysis.} We concluded with a discussion on {\system}’s advantages and limitations. Each interview lasted 35-60 minutes.

\subsubsection{Usability}
Overall, the experts are quite satisfied with {\system}.

\noindent\textbf{Learning curve}.
The experts consider {\system} easy to use.
They like that the matrix view takes a format similar to a data table and naturally supports table operations, such as sorting and swapping.
This familiar design has enhanced their analysis efficiency.
They also indicate that using {\system} enables them to quickly understand the model, which has increased their trust in the model and their confidence in decision-making. \weikai{
Specifically, they expressed satisfaction with
the dynamic hierarchy, which enables them to analyze different subsets of rules on demand.
The experts also value the ease of finding anomalous rules using {\system}.
As E$_2$ commented, ``Different from RuleMatrix~\cite{ming2018rulematrix} and ExplainableMatrix~\cite{neto2020explainable}, this tool automatically recommends some unusual rules, which help me quickly identify potential issues in the model and make interventions accordingly.''}
\lizhenrev{We believe that conducting a user study to compare {\system} with these interactive systems would provide deeper insights into its usability and its effectiveness in facilitating model understanding.
We consider this an important future work.}

\noindent\weikai{\noindent\textbf{Scalability}.
Matrix visualizations often face the scalability issue due to the limited number of rows and columns \jwrevision{that can be displayed}
~\cite{ming2017understanding,yang2022diagnosing}.
To tackle this issue, {\system} 1) organizes a large rule set into a hierarchy and dynamically selects representative rules to display, and 2) sorts attributes by importance and divides them into pages.
These two strategies are effective because \lizhenrev{tree ensemble models typically rely on a smaller set of key attributes for decision making~\cite{breiman2001random}, and} users usually focus on a small subset of \lizhenrev{representative rules and key attributes for initial analysis.}
In addition, our tool can \jwrevision{easily be} 
extended to handle other scalability issues. \jwrevision{For example,} 
for 
attributes with thousands of categories, we can employ alternative feature transformation methods, such as Target Encoding~\cite{geron2022hands}, which replaces each category with a single numerical value representing its statistical correlation to the target variable.
For classification tasks involving more than 10 classes, we can hierarchically organize classes and dynamically assign colors during user exploration ~\cite{chen2024dynamic}.}

\noindent\weikai{\noindent\textbf{Extensibility}.
While {\system} is primarily designed for tree ensemble classifiers, it can be easily extended for regression models.
To achieve this, users simply need to substitute the classification-specific loss with a regression-specific loss during model reduction and employ sequential color schemes to encode the regression values of each rule.
In addition, E$_6$ also pointed out that {\system} can be used to explain other models beyond tree ensemble models: ``Similar to RuleMatrix~\cite{ming2018rulematrix}, this tool supports the analysis of other machine learning models using model surrogate techniques.
This greatly extends the applicability of the tool.''}

\subsubsection{Limitations}

In addition to the positive feedback, the experts also point out several limitations that offer directions for future research.

\noindent\textbf{Online model update}. 
In the current analysis, after the experts identify problems and make corrections, they need to retrain the model and load it back into the system. The experts highlighted that the ability to incrementally update parts of the model online to examine the result would help improve the analysis efficiency. Exploring methods for incrementally updating the model and effectively visualizing the changes presents an interesting research direction.

\noindent\textbf{What-if analysis}.
Another interesting direction is the what-if analysis of rules. E$_6$ suggested allowing examination of how different rule conditions impact the output. However, further exploration is needed on how to use what-if analysis to more accurately guide experts in improving the model.
This research will focus on understanding model rules and making corresponding modifications. We plan to gather more specific requirements from experts to guide future research.

\noindent\textbf{Algorithm efficiency}.
The current efficiency of the model reduction algorithm \weikai{is not as high as desired.
The running time ranges from tens of seconds to a few hours, depending on the model complexity and data size.
For example, it takes around 1 minute for the model in the first case study and takes 3-4 hours in the second case study.}
\lizhen{However, since model reduction is performed offline,  
experts find this time acceptable, as they are willing to \jwrevision{trade off time for scalability}  
in the big data era.
}
%
\weikai{Nevertheless, accelerating this process allows experts to start their analysis at the earliest opportunity.}
The bottlenecks here lie in the grid search for the algorithm parameters and linear programming in Sec.~\ref{sec:reduction}.
In the future, we will design a method for quickly determining algorithm parameters, and simplify the linear programming by retaining more important constraints and variables to accelerate problem solving.

\section{Conclusion}

In this paper, we have presented {\system}, a visual analysis tool that helps domain experts understand the rules extracted from tree ensemble models.
To address the scalability in handling large-scale rule sets, we combine an anomaly-biased model reduction method with a matrix-based hierarchical visualization. 
Organizing rules into a hierarchy achieves scalability without sacrificing fidelity. 
We also propose a dynamic way to construct the rule hierarchy to accommodate different analysis preferences. 
The anomaly-biased model reduction preserves both common and anomalous rules at each level, and thus enables not only an understanding of the overall model behavior, but also the identification of potential flaws in the model. 
A quantitative evaluation and two case studies are conducted to demonstrate the effectiveness of {\system} and its usefulness in real-world applications.

\section*{Acknowledgments}
This work was supported by the National Natural Science Foundation of China under grants U21A20469, 61936002, and in part by Tsinghua-Kuaishou Institute of Future Media Data. The authors would like to thank Xiting Wang and Duan Li for their valuable contributions to the discussions, and Yiwei Hou for her assistance in voicing our video.
\bibliographystyle{abbrv-doi}
\bibliography{reference}

\begin{IEEEbiography}
[{\includegraphics[width=1in,height=1.25in,clip,keepaspectratio]{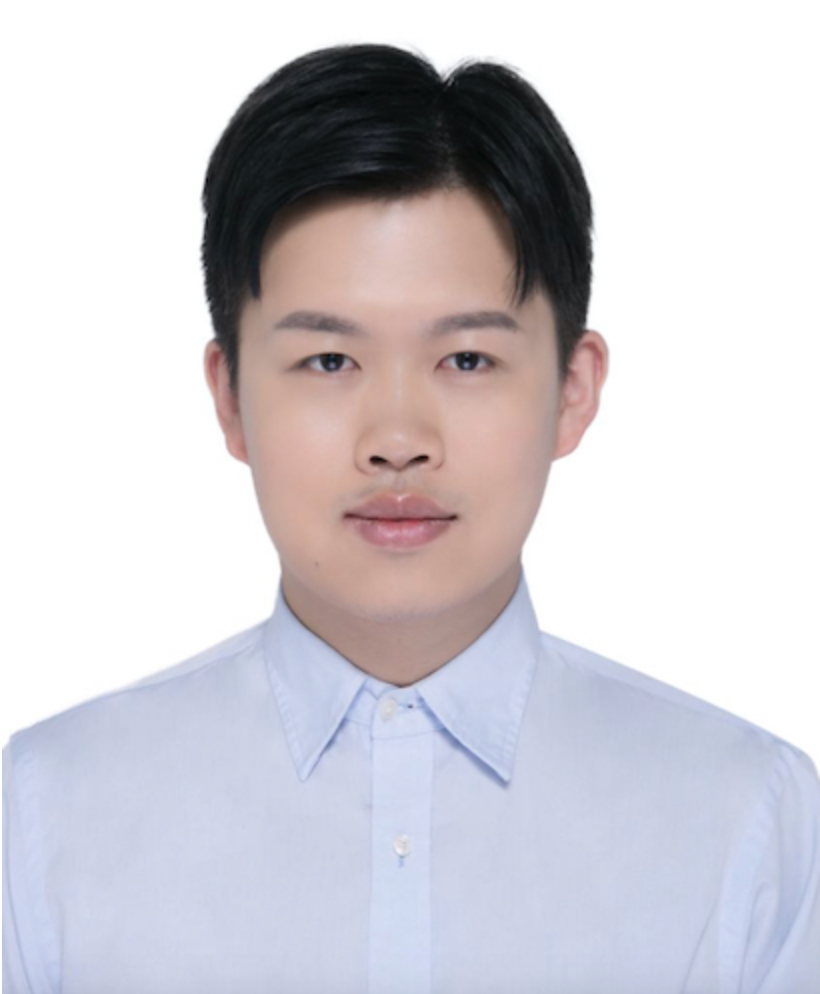}}]
{{Zhen Li}} is a fourth-year Ph.D. student of Software School, Tsinghua University. His research interest is explainable artificial intelligence. He received a B.S. degree from
Tsinghua University and a M.Phil. degree from Hong Kong University of Science and Technology.
\end{IEEEbiography}

\begin{IEEEbiography}
[{\includegraphics[width=1in,height=1.25in,clip,keepaspectratio]{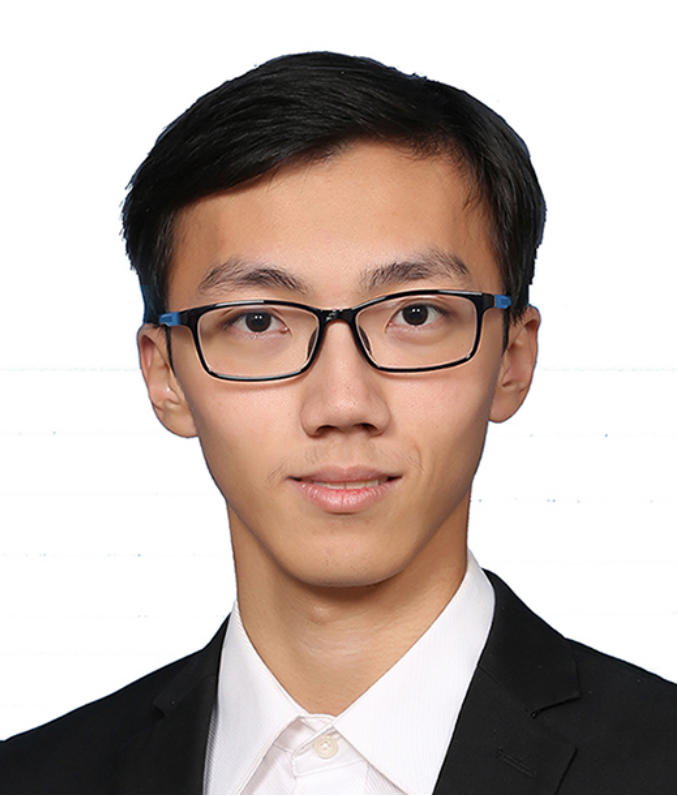}}]{{Weikai Yang}} is an assistant professor in Hong Kong University of Science and Technology (Guangzhou). His research interests lie in visual analytics, machine learning, and data quality improvement. He received a B.S. and a Ph.D from Tsinghua University.\looseness=-1
\end{IEEEbiography}

\begin{IEEEbiography}
[{\includegraphics[width=1in,height=1.25in,clip,keepaspectratio]{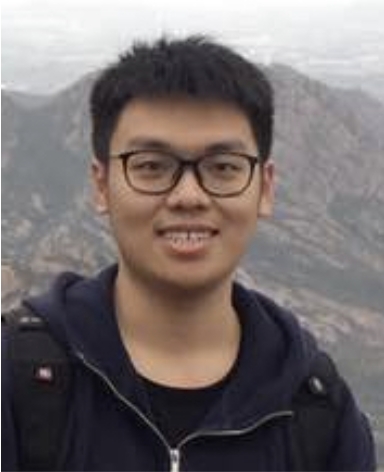}}]
{{Jun Yuan}} is currently a Ph.D. student at Tsinghua University. His research interests are in explainable artificial intelligence. He received a B.S. degree from Tsinghua University.
\end{IEEEbiography}

\begin{IEEEbiography}
[{\includegraphics[width=1in,height=1.25in,clip,keepaspectratio]{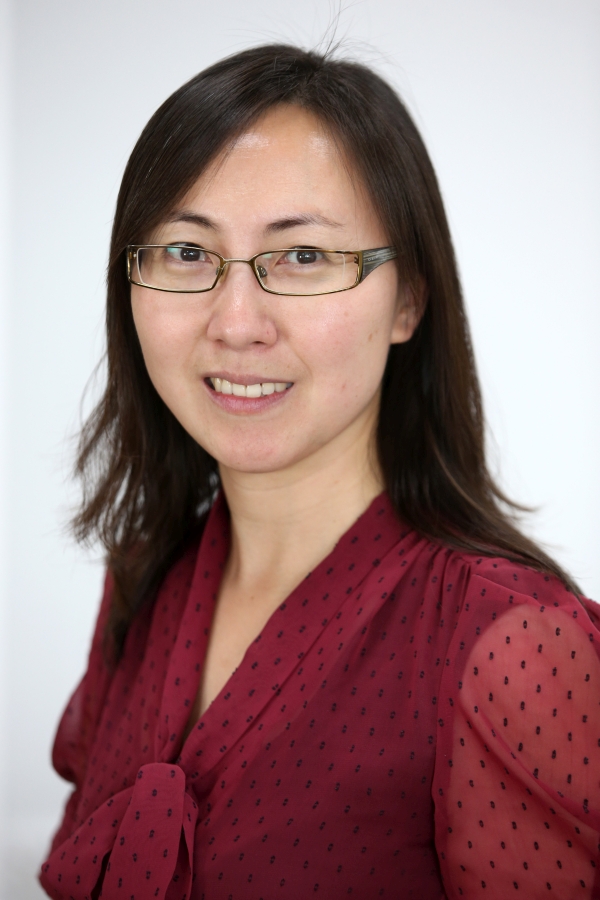}}]{Jing Wu} is a lecturer in computer science and informatics at Cardiff University, UK. Her research interests are in computer vision and graphics including image-based 3D reconstruction, face recognition, machine learning and visual analytics. She received BSc and MSc from Nanjing University, and Ph.D. from the University of York, UK. She serves as a PC member in CGVC, BMVC, etc.
\end{IEEEbiography}

\begin{IEEEbiography}[{\includegraphics[width=1in, height=1.25in, clip, keepaspectratio]{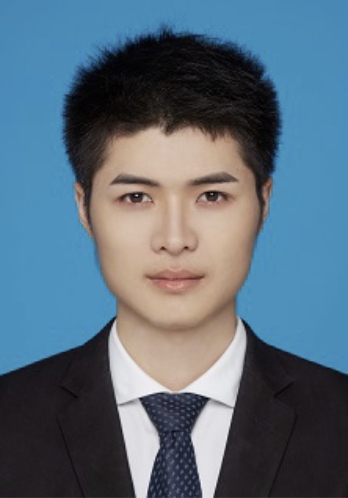}}]{Changjian Chen} is an assistant professor at Hunan University. He received a Ph.D. from Tsinghua University and a B.S. from University of Science and Technology of China. His research interests focus on visual analytics and machine learning, especially visual analysis methods to improve training data quality.
\end{IEEEbiography}

\begin{IEEEbiography}[{\includegraphics[width=1in, height=1.25in, clip, keepaspectratio]{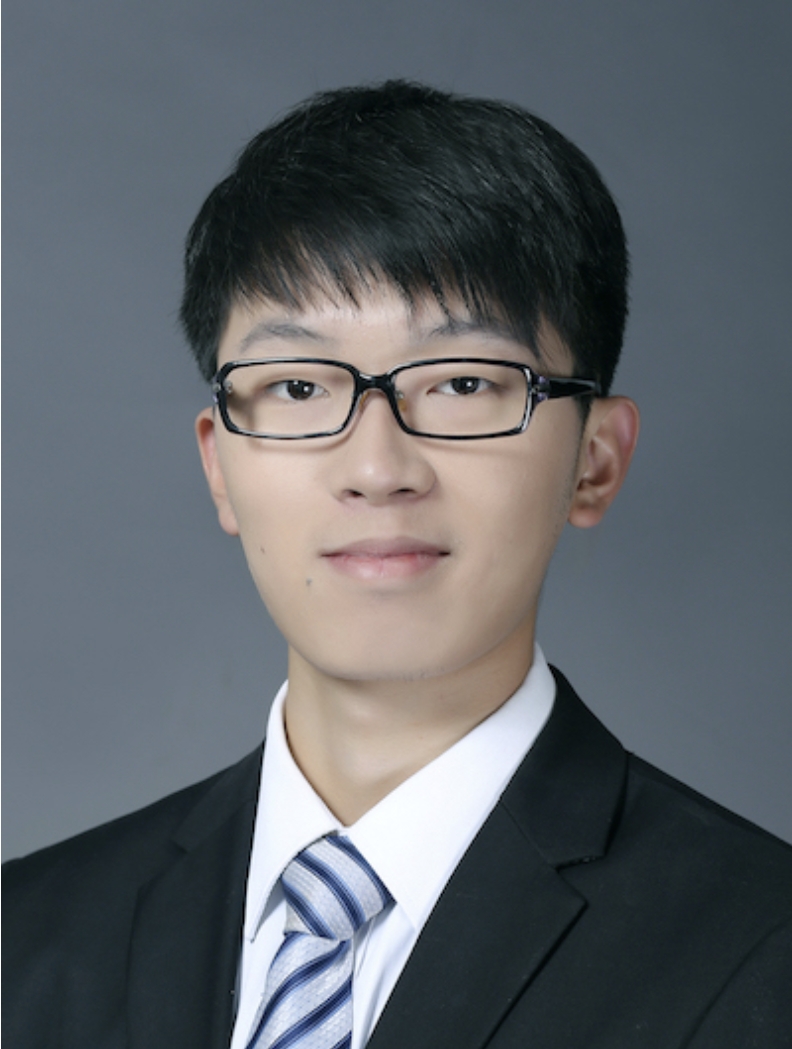}}]{{Yao Ming}} is a quantitative researcher at Citadel Securities. His research focus on visual analytics, explainable machine learning, and natural language processing. He received a Ph.D. in Computer Science from the Hong Kong University of Science and Technology and a B.S. from Tsinghua University. For more details please refer
to https://www.myaooo.com.
\end{IEEEbiography}

\begin{IEEEbiography}
[{\includegraphics[width=1in,height=1.25in,clip,keepaspectratio]{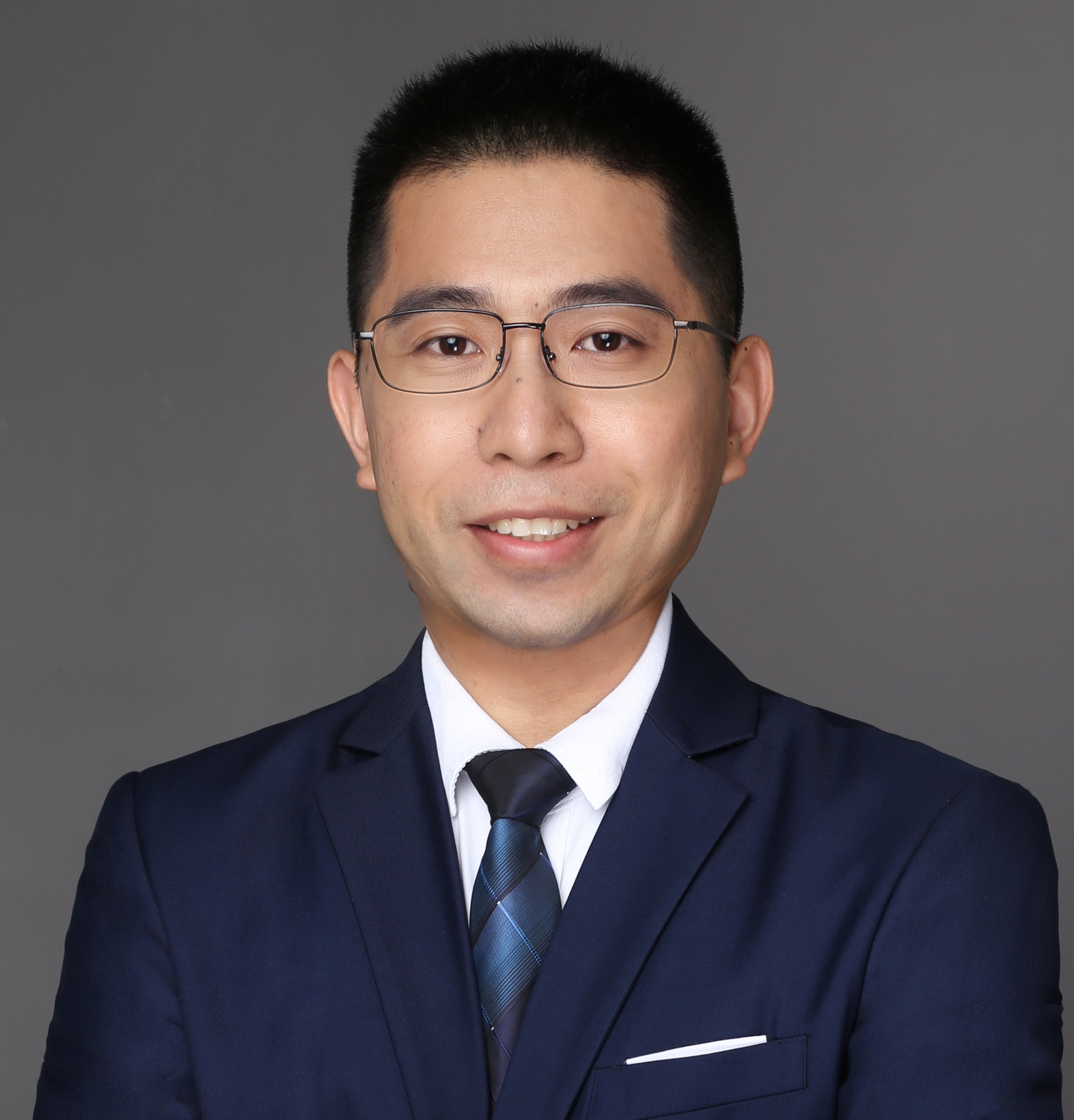}}]{Fan Yang} is an algorithm engineer at Kuaishou. His research focuses on video understanding, large language model, and joint modeling of content and behavior. He received a M.S. from Tsinghua University and B.S. from Beijing University of Posts and Telecommunications.
\end{IEEEbiography}

\begin{IEEEbiography}
[{\includegraphics[width=1in,height=1.25in,clip,keepaspectratio]{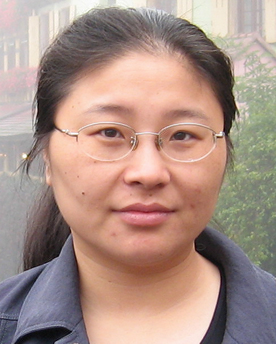}}]{Hui Zhang}
is an Associate Professor at School of Software, Tsinghua University, China. She received her B.Sc. and Ph.D. in Computer Science from Tsinghua University, in 1997 and 2003, respectively. Her research interests include computer aided design and computer graphics.
\end{IEEEbiography}

\begin{IEEEbiography}
[{\includegraphics[width=1in,height=1.25in,clip,keepaspectratio]{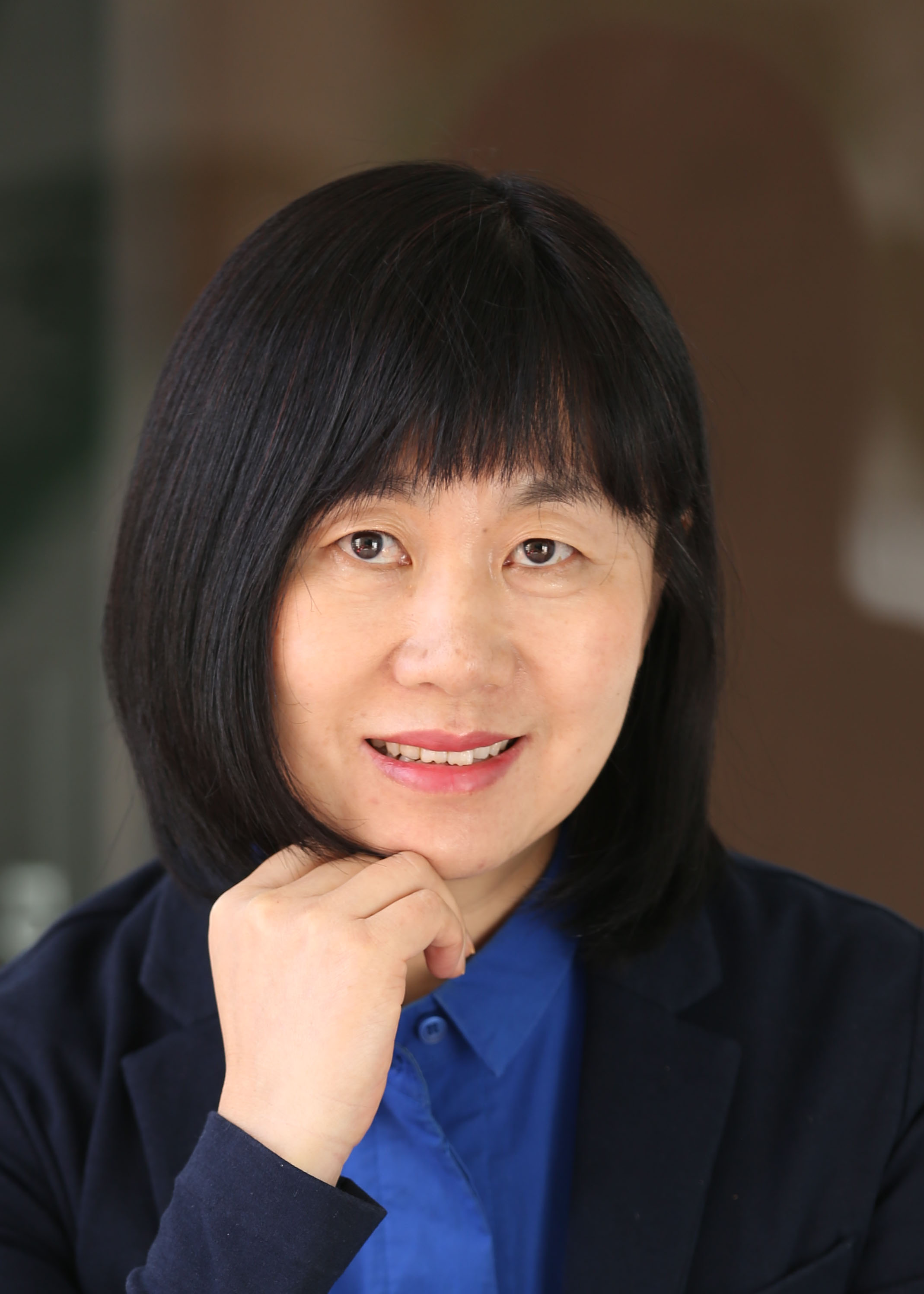}}]{Shixia Liu}
is a professor at Tsinghua University. Her research interests include explainable artificial intelligence, visual analytics for big data. She worked as a research staff member at IBM China Research Lab and a lead researcher at Microsoft Research Asia.
She received a B.S. and M.S. from Harbin Institute of Technology, a Ph.D. from Tsinghua University.
She is a fellow of IEEE and an associate editor-in-chief of IEEE Trans. Vis. Comput. Graph.
\end{IEEEbiography}

\vfill

\end{document}